\documentclass[11pt]{article}

\usepackage[english]{babel}
\usepackage[utf8x]{inputenc}
\usepackage[T1]{fontenc}
\usepackage{scalerel}

\usepackage[a4paper,top=3cm,bottom=2cm,left=3cm,right=3cm,marginparwidth=1.75cm]{geometry}

\usepackage{booktabs}       
\usepackage{amsfonts}       
\usepackage{nicefrac}       
\usepackage{microtype}      

\usepackage{url}            
\usepackage{amsmath}
\usepackage{color}
\usepackage{amssymb}
\usepackage{amsthm}
\usepackage{pifont}
\usepackage{algorithm}
\usepackage{algpseudocode}

\usepackage{paralist}
\usepackage{dsfont}
\usepackage{natbib}
\usepackage{csquotes}

\usepackage{amsmath}
\usepackage{amssymb}
\usepackage{amsthm}
\usepackage{mathtools}

\usepackage{microtype}
\usepackage{graphicx}
\usepackage{booktabs} 
\usepackage{dirtytalk}
\usepackage{subfiles}

\usepackage{xspace}
\usepackage{forloop}
\usepackage{multirow}

\usepackage[pdftex,dvipsnames,table]{xcolor}
\usepackage[colorinlistoftodos,prependcaption,textsize=small]{todonotes}
\usepackage{graphicx}
\usepackage[shortlabels]{enumitem}
\usepackage{bbm}

\usepackage[colorlinks=true,linkcolor=blue,citecolor=blue]{hyperref}

\title{Task-Robust Model-Agnostic Meta-Learning}
\author{Liam Collins\thanks{Department of Electrical and Computer Engineering, 
The University of Texas at Austin, Austin, TX,  USA. \{Email: liamc@utexas.edu, mokhtari@austin.utexas.edu, sanjay.shakkottai@utexas.edu\}.} , Aryan Mokhtari$^*$, Sanjay Shakkottai$^*$}

\begin{document}

\maketitle

\begin{abstract}
Meta-learning methods have shown an impressive ability to train models that rapidly learn new tasks. However, these methods only aim to perform well in expectation over tasks coming from some particular distribution that is typically equivalent across meta-training and meta-testing, rather than considering worst-case task performance. In this work we introduce the notion of ``task-robustness'' by reformulating the popular Model-Agnostic Meta-Learning (MAML) objective \citep{finn2017model} such that the goal is to minimize the maximum loss over the observed meta-training tasks. The solution to this novel formulation is task-robust in the sense that it places equal importance on even the most difficult and/or rare tasks. This also means that it performs well over all distributions of the observed tasks, making it robust to shifts in the task distribution between meta-training and meta-testing.
We present an algorithm to solve the proposed min-max problem, and show that it converges to an $\epsilon$-accurate point at the optimal rate of $\mathcal{O}(1/\epsilon^2)$ in the convex setting and to an $(\epsilon, \delta)$-stationary point at the rate of $\mathcal{O}(\max\{1/\epsilon^5, 1/\delta^5\})$ in nonconvex settings. We also provide an upper bound on the new task generalization error that captures the advantage of minimizing the worst-case task loss, and demonstrate this advantage in sinusoid regression and image classification experiments.
\end{abstract}
\newcommand{\ones}{\mathbf 1}
\newcommand{\integers}{{\mbox{\bf Z}}}
\newcommand{\symm}{{\mbox{\bf S}}}  

\newcommand{\nullspace}{{\mathcal N}}
\newcommand{\range}{{\mathcal R}}
\newcommand{\Rank}{\mathop{\bf Rank}}
\newcommand{\Tr}{\mathop{\bf Tr}}
\newcommand{\diag}{\mathop{\bf diag}}
\newcommand{\card}{\mathop{\bf card}}
\newcommand{\rank}{\mathop{\bf rank}}
\newcommand{\conv}{\mathop{\bf conv}}
\newcommand{\prox}{\mathbf{prox}}

\newcommand{\ind}{\mathds{1}}
\newcommand{\E}{\mathbb{E}}
\newcommand{\Prob}{\mathbb{P}}
\newcommand{\bigO}{\mathcal{O}}
\newcommand{\B}{\mathcal{B}}
\newcommand{\s}{\mathcal{S}}
\newcommand{\Ev}{\mathcal{E}}
\newcommand{\R}{\mathbb{R}}
\newcommand{\Co}{{\mathop {\bf Co}}} 
\newcommand{\dist}{\mathop{\bf dist{}}}
\newcommand{\argmin}{\mathop{\rm argmin}}
\newcommand{\argmax}{\mathop{\rm argmax}}
\newcommand{\epi}{\mathop{\bf epi}} 
\newcommand{\Vol}{\mathop{\bf vol}}
\newcommand{\dom}{\mathop{\bf dom}} 
\newcommand{\intr}{\mathop{\bf int}}
\newcommand{\sign}{\mathop{\bf sign}}

\newcommand{\cf}{{\it cf.}}
\newcommand{\eg}{{\it e.g.}}
\newcommand{\ie}{{\it i.e.}}
\newcommand{\etc}{{\it etc.}}

\newtheorem{theorem}{Theorem}
\newtheorem{remark}{Remark}
\newtheorem{definition}{Definition}
\newtheorem{corollary}{Corollary}
\newtheorem{proposition}{Proposition}
\newtheorem{lemma}{Lemma}
\newtheorem{fact}{Fact}
\newtheorem{assumption}{Assumption}

\newcommand{\numberthis}{\addtocounter{equation}{1}\tag{\theequation}}

\newcommand{\simiid}{\overset{\text{i.i.d.}}{\sim}}

\section{Introduction}

Despite continual advances in computational power and data collection, many scenarios remain in which machine learning models must rapidly adapt to previously unseen tasks. Motivated by such scenarios, meta-learning techniques aim to learn how to learn quickly from few samples by leveraging knowledge acquired while learning prior tasks  \citep{bengio1990learning, thrun2012learning}. The recent successes of these techniques in areas such as few-shot learning \citep{finn2017model,  ravi2016optimization, snell2017prototypical, vinyals2016matching} and reinforcement learning \citep{duan2016rl, song2019esmaml, wang2016learning} have sparked tremendous interest in meta-learning.

Following the setting introduced by \citet{baxter1998theoretical}, most offline meta-learning methods 
try to minimize the expected loss on new tasks drawn from the same, but unknown, distribution as a finite set of meta-training tasks. For example, in gradient-based meta-learning, the learning method is typically a small number of stochastic gradient descent (SGD) steps, and the means to learn quickly is having a favorable initialization. Standard methods thus try to find an initialization that enables the model fine-tuned via task-specific SGD to perform well in expectation over new tasks. Since they assume the new tasks are drawn from the same unknown distribution as the meta-training tasks, during meta-training they attempt to minimize the average empirical loss after one step of SGD \citep{finn2017model, nichol2018reptile}.

However, by minimizing the average loss, such methods may perform arbitrarily poorly on difficult and/or rare meta-training tasks. Poor worst-case performance is unacceptable in applications including those where safety and fairness are critical, e.g., few-shot fingerprint recognition in security systems and few-shot facial recognition among different demographics.
Moreover, the assumption that the meta-training and meta-testing distributions are equivalent is often unrealistic.
If the meta-training dataset overestimates the prevalence of certain types of tasks in the meta-test distribution, existing methods will overfit to the popular tasks and fail to generalize to new tasks in both expectation and the worst case.
Indeed, existing generalization bounds for gradient-based meta-learning strategies depend on the similarity of the meta-test tasks to the meta-training solution \citep{zhou2019efficient, khodak2019provable},
rather than exploiting the diversity of the meta-training tasks to show generalization to a broad range of new tasks.

To address these issues, we propose a novel meta-learning formulation that calls for minimizing the maximum as opposed to average task loss during meta-training. 
 Our contributions are threefold: 
\begin{itemize}
\item We modify the standard gradient-based meta-learning framework, Model-Agnostic Meta-Learning (MAML) \citep{finn2017model}, to find an initialization that minimizes the loss after one SGD step for the \textit{worst-case} task, where tasks are broadly defined as distributions over few-shot learning problems.
Our new formulation, Task-Robust MAML (TR-MAML), thus yields a "task-robust" solution, in the sense that it prioritizes performance equally on all observed tasks, including the hardest and rarest ones. Importantly, this means it is also robust to all shifts in distribution over the sampled tasks from meta-training to meta-testing.
\item We present an algorithm to solve our min-max formulation and prove that it convergences efficiently in both convex and nonconvex settings. In the convex case, it achieves the optimal rate of $\mathcal{O}(\epsilon^{-2})$ stochastic gradient evaluations, and in the nonconvex case, it reaches an $(\epsilon, \delta)$-stationary point at a rate of $\mathcal{O}(\max \{ \epsilon^{-5}, \delta^{-5}\} )$ stochastic gradient evaluations.
\item We capture the generality of our formulation's task robustness by giving a Rademacher complexity bound on the generalization error of any new task within the convex hull of the meta-training tasks, as well as showing improved performance in few-shot sinusoid regression and image classification experiments compared to MAML.
\end{itemize}

\noindent \textbf{Related Work.} 
Among a variety of meta-learning formulations, MAML \citep{finn2017model} has become especially popular due to its efficiency and flexibility, inspiring many related works
\citep{antoniou2018train, li2017meta, nichol2018reptile, bertinetto2018meta, lee2019meta}. From more theoretical perspectives,
 \citet{fallah2019convergence} analyzed the convergence of MAML with nonconvex losses, \citet{rajeswaran2019meta} and \citet{zhou2019efficient} presented MAML variants with guarantees in both convex and nonconvex settings, and other works have shown regret bounds for online analogues of MAML \citep{finn2019online, zhuang2019online, khodak2019adaptive}.
Meanwhile, robustness in meta-learning has been studied in multiple recent works. \citet{zugner2019adversarial} and \citet{yin2018adversarial} proposed models whose expected performance is robust to perturbations in the task samples,
and \citet{Lee2020Learning} extended MAML to deal with imbalances in the number of samples per task instance and out-of-distribution meta-test tasks, but their model requires a complicated dataset encoding and computing per-task balancing variables. Additionally, \citet{Jamal_2019} introduced a heuristic that aims to prevent over-performing on certain meta-training tasks by regularizing the inequality among task losses, although only across mini-batches. \citet{cai2020weighted} also considered a task-weighted objective and showed Rademacher complexity-based generalization bounds, but their weights utilize task similarity to a particular target rather than optimizing for worst-case performance.
To the best of our knowledge, no other offline meta-learning formulation attempts to minimize the worst-case loss over tasks.

Many works outside meta-learning have considered min-max optimization problems of the finite-sum form discussed here. In the context of distributionally-robust optimization, \citet{shalev2016minimizing} and \citet{duchi2018learning}
argued that minimizing the maximal loss over a set of possible distributions can provide better generalization performance than minimizing the average loss.
While \citet{nemirovski2009robust} showed that the stochastic mirror descent-ascent algorithm achieves 
the asymptotically optimal $\mathcal{O}(\epsilon^{-2})$ convergence rate to an $\epsilon$-accurate solution in the convex setting, the literature is less established for nonconvex problems. 
\citet{rafique2018non} proposed a stochastic inexact proximal point method that attains $\tilde{\mathcal{O}}(\epsilon^{-6})$ convergence in terms of the outer minimization problem when that problem is nonsmooth and weakly convex, while \citet{qian2019robust} showed $\tilde{\mathcal{O}}(\epsilon^{-4})$ convergence when the outer problem is smooth and strongly convex. In the deterministic case, \citet{nouiehed2019solving} demonstrated an $\tilde{\mathcal{O}}(\epsilon^{-3.5})$ convergence rate to an $\epsilon$-first-order Nash equilibrium for a gradient descent-ascent algorithm.
Also, \citet{chen2017robust} and \citet{jin2019minmax} analyzed first order methods that improve on these rates but rely on an oracle to solve the inner maximization. 

\section{Problem Formulation}\label{sec:problem_formulation}

Before discussing our min-max objective, we first formalize the meta-learning scenario. Let $x \in \mathcal{X}$ and $y \in \mathcal{Y}$ denote inputs and labels, respectively, and let $h_w : \mathcal{X} \rightarrow \mathcal{Y}$ represent the model parameterized by $w$. The performance of $h_w$ on a point $(x,y) \in \mathcal{X} \times \mathcal{Y}$ is determined by $\ell(h_w(x), y)$, where $\ell: \mathcal{Y} \times \mathcal{Y} \rightarrow \mathbb{R}_+$ is a loss function, e.g., the mean squared error in regression and the cross entropy loss in classification. We define a task $\mathcal{T}_i$ as a distribution $\mathcal{D}_i$ over task instances, which are few-shot learning episodes composed of two data batches, $D^{\text{train}}_{i,j}$ and $D^{\text{test}}_{i,j}$, of $K$ and $J$ points, respectively, in $\mathcal{X} \times \mathcal{Y}$. Within each task instance, the goal of the learner is to perform well on the points in $D^{\text{test}}_{i,j}$ after learning from the points in $D^{\text{train}}_{i,j}$, which is made possible by assuming that each point in both batches is an i.i.d. sample from the same distribution $\mathcal{D}_{i,j}$ over $\mathcal{X} \times \mathcal{Y}$.

During meta-training, a finite number of task instances are
observed by first sampling a task $\mathcal{T}_i$ from $P(\mathcal{T})$, the meta-training distribution over tasks, then sampling $(D^{\text{train}}_{i,j}, D^{\text{test}}_{i,j}) \sim \mathcal{D}_i$. Let there be $m_i$ instances of the $i$-th task for each of $n$ tasks observed during meta-training, for a total of $m \coloneqq \sum_{i=1}^n m_i$ task instances. 
In MAML, for each task instance, the dataset $D^{\text{train}}_{i,j}$ is used to update a global initialization $w$ via one SGD step with respect to the expected loss of the model on $\mathcal{D}_{i,j}$, namely
$f_{i,j}(w) \coloneqq \mathbb{E}_{(x,y)\sim \mathcal{D}_{i,j}} [ \ell(h_w(x), y)]$.
Afterwards, the resulting "test" loss is approximated using $D^{\text{test}}_{i,j}$, which serves as the meta-training loss.
With the ultimate goal of learning how to learn new task instances coming from the same distribution $P(\mathcal{T})$, the meta-training objective is to find a $w$ that minimizes the post-update loss on $D^{\text{test}}_{i,j}$ on average over the observed task instances, namely:
\begin{equation} \label{typical_meta_sample_avg}
    \min_{w \in \mathcal{W}}  \frac{1}{m} \sum_{i=1}^n  \sum_{j=1}^{m_i}\hat{f}_{i,j}(w - \alpha \nabla \hat{f}_{i,j}(w; D_{i,j}^{\text{train}}), D^{\text{test}}_{i,j}),
\end{equation} 
where $\alpha$ is the inner update step size and $\hat{f}_{i,j}(\cdot, D^{\text{test}}_{i,j}) = \frac{1}{J} \sum_{(x,y) \in D^{\text{test}}_{i,j}}  \ell(h_w(x), y)$ is the sample-average approximation of $f_{i,j}(\cdot)$ using the $J$ samples in $D^{\text{test}}_{i,j}$, and likewise for $\nabla \hat{f}_{i,j}(\cdot, D^{\text{train}}_{i,j})$.
As referred to in the introduction, the solution of \eqref{typical_meta_sample_avg} may perform arbitrarily poorly on tasks that differ significantly from the average task instance,  
which is especially problematic
if tasks similar to those become more prevalent at meta-test time due to a distributional shift.
Thus, we propose to treat all $n$ meta-training tasks equally by minimizing the maximum task empirical average meta-loss 
$\hat{F}_i(w)$:
\vspace{0mm}
\begin{equation} \label{top-maxi_emp}
	\min_{w \in \mathcal{W}} \max_{i \in [n]} \bigg\{ \hat{F}_i(w) \coloneqq \frac{1}{m_i} \sum_{j=1}^{m_i}   \hat{f}_{i,j}  (w - \alpha \nabla \hat{f}_{i,j} (w, D_{i,j}^{\text{train}}), D_{i,j}^{\text{test}}) \bigg\}.
\end{equation}
Problem \eqref{top-maxi_emp} is equivalent to the problem of finding the $w^{\ast}$ that minimizes the worst-case meta-learning performance over all distributions of the $n$ tasks, since the worst-case distributions will occur at the extreme points of the probability simplex in $n$ dimensions. We write this relaxed problem as
\begin{equation} \label{top-relaxed}
   \min_{w \in \mathcal{W}} \max_{p \in \Delta_n} \bigg\{\phi(w,p) \coloneqq \sum_{i=1}^n p_i \hat{F}_i(w)\bigg\},
\end{equation}
where $p_i$ is the probability associated with task $i$, the vector $p = (p_1,\dots, p_n)$ is the concatenation of probabilities, and 
$\Delta_n = \{ p \in \mathbb{R}^n_+ \!\mid\! \sum_{i=1}^n p_i \!=\! 1\}$. 
Note that \eqref{top-relaxed} may be hard to solve if $n$ is very large, and in many applications, $m$ is indeed very large. However, $n$ need not be, as tasks may be defined to encompass many similar task instances.
We provide experiments for this case in Section~\ref{section:experiments}.

By optimizing for worst-case performance, the formulation in \eqref{top-relaxed} encourages a solution $w^\ast$ that performs similarly across all of the observed tasks. Instead of disregarding performance on some tasks, any algorithm that solves \eqref{top-relaxed} must try to perform reasonably well on all of them. Indeed, as observed in \citep{duchi2016statistics}, the min-max formulation implicitly regularizes the variance of the losses. This naturally makes the solution robust to distributional shifts between meta-training and meta-testing, and we provably show its ability to generalize to new tasks in Section \ref{sec:generalization}.

\vspace{0mm}
\section{Algorithm}\label{sec:algorithm}
\vspace{0mm}
Taking inspiration from \citep{nemirovski2009robust}, we propose to solve the meta-training problem \eqref{top-relaxed} using 
a Euclidean version of the robust stochastic mirror-prox algorithm. Our method, termed TR-MAML and outlined in Algorithm \ref{alg1}, requires stochastic gradient estimates of the function $\phi(w,p)$ defined in \eqref{top-relaxed} with respect to $w$ and $p$. Note that the full gradients, denoted by $g_w(w,p)$ and $g_p(w,p)$, respectively, are
\begin{align}
    g_w(w,p) &= \sum_{i=1}^n \frac{p_i}{m_i} \sum_{j=1}^{m_i} (I - \alpha \nabla^2 \hat{f}_{i,j}(w, D_{i,j}^{\text{train}})) \nabla \hat{f}_{i,j}(w - \alpha \nabla \hat{f}_{i,j}(w, D_{i,j}^{\text{train}}), D_{i,j}^{\text{test}}), \label{g_1} \\
    g_p(w,p) &= \bigg[ \frac{1}{m_i} \sum_{j=1}^{m_i}   \hat{f}_{i,j}  (w - \alpha \nabla \hat{f}_{i,j} (w, D_{i,j}^{\text{train}}), D_{i,j}^{\text{test}}) \bigg]_{1 \leq i \leq n} \label{g_2},
\end{align}
where $\nabla^2 \hat{f}_{i,j}(w,D^\text{train}_{i,j})$ is the sample average approximation of $\nabla^2 f_{i,j}(w)$ based on the $K$ samples in $D^\text{train}_{i,j}$, and the notation $[a_i]_{1\leq i \leq n}$ corresponds to the vector $[a_1,\dots,a_n] \in \mathbb{R}^n$. 
Since $n$ and the $m_i$'s may be large, TR-MAML must estimate the full gradients $g_w$ and $g_p$ on each iteration.
To do so, it first uniformly and independently samples a set $\mathcal{C}$ of $C$ indices $\{i_k\}_{k=1}^C$ from $\{1,\dots,n\}$. For each $i_k \in \mathcal{C}$, the algorithm samples one index $j_k$ uniformly from $\{1, \dots, m_i\}$, then estimates $g_w(w,p)$ and $g_p(w,p)$ using the data $\{(D_{i_k,j_k}^{\text{train}}, D_{i_k,j_k}^{\text{test}})\}_{k=1}^C$.
The two estimates can then be written as
\vspace{0mm}
\begin{align}
\hat{g}_w(w,p) &= \frac{n}{C} \sum_{k=1}^C p_{i_k}  (I \!-\! \alpha \nabla^2 \hat{f}_{i_k, j_k}(w, D_{i_k,j_k}^{\text{train}})) \nabla \hat{f}_{i_k, j_k}(w - \alpha \nabla \hat{f}_{i_k, j_k}(w, D_{i_k,j_k}^{\text{train}}), D_{i_k,j_k}^{\text{test}}) ,
\label{gradw_comp} \\
\hat{g}_p(w,p) &=  \frac{n}{C} \sum_{k=1}^C \hat{f}_{i_k, j_k} (w -  \alpha \nabla \hat{f}_{i_k, j_k}(w,D^{\text{train}}_{i_k,j_k}), D^{\text{test}}_{i_k,j_k}) e_{i_k}, \
\label{gradp_comp}
\end{align}
where $e_{i_k}$ is the $i_k$-th standard basis vector in $\mathbb{R}^n$. We show that $\hat{g}_w(w,p)$ and $\hat{g}_p(w,p)$ are unbiased and bound their second moments in Section~\ref{sec:theory}.
In order to solve \eqref{top-relaxed}, TR-MAML initializes $p^0 = \left[ {1}/{n} \right]_{1\leq i \leq n}$ and $w^0 \in \mathcal{W}$, then executes alternating projected stochastic gradient descent-ascent. 
In particular, from iterations $t = 0$ to $T-1$, TR-MAML computes
$w^{t+1}$ and $p^{t+1}$ as
\begin{align}
    w^{t+1} = \Pi_{\mathcal{W}}(w^{t} - \eta^{t}_w \hat{g}_w(w^t, p^t)), \; \quad p^{t+1} = \Pi_{\Delta_n} (p^{t} + \eta^{t}_p \hat{g}_p(w^t, p^t)),\label{updatew}
\end{align}
where $\eta^{t}_w$ and $ \eta^{t}_p$ are step sizes, $\Pi_W(u) = \arg \min_{w \in \mathcal{W}} \|u-w\|_2$ and $\Pi_{\Delta_n}(q) = \arg \min_{p \in \Delta_n} \|p-q\|_2$. 
The projections are convex programs and can be solved efficiently using standard techniques. 
In particular, since $\Delta_n$ is the full simplex, $\Pi_{\Delta_n}(q)$ can be computed in $\mathcal{O}(n \log n)$ time \citep{wang2013projection}. As mentioned previously, tasks can be defined to leverage similarity among the task instances such that $n$ is small, in which case the $\mathcal{O}({d^2})$ per-iteration cost of both MAML and TR-MAML due to the Hessian estimations trivializes the added cost of the simplex projection in TR-MAML, thus TR-MAML has effectively the same computational cost as MAML. Nevertheless, first-order MAML approximations \citep{finn2017model, nichol2018reptile, fallah2019convergence} may be seamlessly applied to TR-MAML to reduce the computational burden.
After $T$ iterations, TR-MAML terminates in one of two ways:
\noindent \textbf{Case T1.} If each 
$\hat{F}_i(w)$ is convex, TR-MAML outputs 
    ${w^{\mathbf{c}}_T \coloneqq \frac{1}{T}\sum_{t=1}^T w^t}$ and $ {p^\mathbf{c}_T \coloneqq \frac{1}{T}\sum_{t=1}^T p^t}$.
\noindent \textbf{Case T2.} Otherwise, TR-MAML samples $\tau$ uniformly from $\{1,...,T\}$ and outputs $w^\tau_T \coloneqq w^\tau $ and $p^\tau_T \coloneqq p^\tau$.
\begin{algorithm}[tb]
   \caption{Task-Robust MAML (TR-MAML)}
   \label{alg1}
\begin{algorithmic}
   \State {\bfseries Input:} $m$ task instances of $n$ unique tasks; parameters $\alpha$, $\{\eta_w^t\}_t$, $\{\eta_p^t\}_t$, $T$,$C$
   \State Initialize $p^1 = [\tfrac{1}{n}]_{1\leq i \leq n}$ and $w^1 \in \mathcal{W}$ arbitrarily.
   \For{$t=0$ {\bfseries to} $T-1$}
   \State Sample a batch $\mathcal{C}$ of $C$ unique task indices uniformly from $\{1,\dots,n\}$.
  \For {$i_k \in \mathcal{C}$}
   \State Sample one task instance index $j_k$ uniformly from $\{1,\dots,m_{i_k}\}$.
   \EndFor
   \State Compute $\hat{g}_w(w^t,p^t)$ and $\hat{g}_p(w^t,p^t)$ using \eqref{gradw_comp} and \eqref{gradp_comp}, respectively.
   \State Update $w^{t+1}$ and $p^{t+1}$ as in \eqref{updatew}.
   \EndFor
   \State {\bfseries Output:} See Cases T1 and T2.
\end{algorithmic}
\end{algorithm}

\vspace{0mm}
\section{Convergence Analysis}\label{sec:theory}
\vspace{0mm}
We next analyze the convergence of TR-MAML to a solution of \eqref{top-relaxed}.
Convergence results for stochastic gradient-based algorithms 
typically assume access to unbiased stochastic gradients with bounded second moments \citep{nemirovski2009robust, rafique2018non}. In our case, $\hat{g}_w$ and $\hat{g}_p$ are naturally unbiased, but bounding their second moments requires modest assumptions on the functions $\hat{f}_{i,j}$ due to the nested structure of $\hat{F}_i$.
\begin{assumption} \label{bound_grad} $\hat{f}_{i,j}(\cdot, D^{\text{train}}_{i,j})$ and $\hat{f}_{i,j}(\cdot, D^{\text{test}}_{i,j})$, $ \forall i \in [n]$ and $j \in [m_i]$ are $\hat{B}$-bounded and $\hat{L}$-Lipschitz. Furthermore, $\lambda_{\min}(\nabla^2 \hat{f}_{i,j}(w,D^{\text{train}}_{i,j})) \geq -\hat{H}$ for all $w \in \mathcal{W}$.
\end{assumption}
\vspace{0mm}
With this assumption, we can bound the second moments. All proofs are given in the appendix.
\vspace{1mm}
\begin{lemma} \label{lemma_unbiased} Under Assumption \ref{bound_grad}, for all $w \in \mathcal{W}, p \in \Delta_n$, vectors $\hat{g}_w(w,p)$ and $\hat{g}_p(w,p)$ satisfy:
(i)~$ \mathbb{E}[\hat{g}_w(w,p) ] =g_w(w,p), \mathbb{E}[\hat{g}_p(w,p) ] = g_p(w,p)$; and (ii) Bounded second moment: $ \mathbb{E}[ \|\hat{g}_w \|_2^2 ] \leq {n}(1+\alpha \hat{H})^2\hat{L}^2; \; \: \mathbb{E}[ \|\hat{g}_p\|_2^2 ] \leq \frac{n(n+C+1)\hat{B}^2}{C}=: \hat{G}^2_p$
\end{lemma}

\noindent \textbf{Convex Setting.} Our first convergence result holds in the case when each $\hat{F}_i$ is convex. Note that the convexity of each $f_{i,j}$ does not imply the convexity of $\hat{F}_i$ (consider as a counterexample $f_{i,j}(w) = 1/w$ for $w \in \mathbb{R}_+ \setminus \{0\}$). In Lemma \ref{lemma-strong} we adapt a result from \citep{finn2019online} showing that the strong convexity of each $\hat{f}_{i,j}(\cdot, D^{\text{test}}_{i,j} )$ implies the strong convexity of $\hat{F}_i$ under an additional assumption on each $\hat{f}_{i,j}(\cdot, D^{\text{train}}_{i,j} )$.
 \begin{assumption} \label{assump_hess}
   $\hat{f}_{i,j}(\cdot, D^{\text{train}}_{i,j})$, for all $j \in [m_i]$, is $\hat{M}$-smooth and $\hat{\rho}$-Hessian-Lipschitz.
\end{assumption}
\begin{lemma} (Adapted from \citep{finn2019online}, Theorem 1)\label{lemma-strong}
Suppose $\alpha < 1/\hat{M}$ and Assumptions \ref{bound_grad} and \ref{assump_hess} hold. If $\hat{f}_{i,j}(\cdot, D^{\text{test}}_{i,j} )$ is $\hat{\mu}$-strongly convex $\forall j \in [m_i]$, then $\hat{F}_i$ is $\tilde{\mu} \coloneqq (\hat{\mu}(1 - \alpha \hat{M})^2 - \alpha \hat{L}\hat{\rho})$-strongly convex.
\end{lemma}

The optimal rate of convergence for solving convex-concave stochastic min-max problems is $\mathcal{O}(1/\epsilon^{2})$, where convergence rate is measured in terms of the expected number of stochastic gradient computations required to achieve a duality gap of $\epsilon$ \citep{nemirovski2009robust}. The duality gap of the pair $(\tilde{w}, \tilde{p})$ is defined as
    $\max_{p \in \Delta_n} \phi(\tilde{w}, p) -\min_{w \in \mathcal{W}} \phi(w,\tilde{p})$. By strong duality, $(\tilde{w}, \tilde{p})$ is optimal if and only if it has a duality gap of zero.
We show that TR-MAML achieves the optimal $\mathcal{O}(1/\epsilon^{2})$ rate by adapting Theorem 2 from \citep{mohri2019agnostic}, which in turn is a simplified version of Theorem 1 from \citep{juditsky2011solving}.

\begin{theorem} (Adapted from \citep{mohri2019agnostic}, Theorem 2) \label{theorem_con}
Consider problem \eqref{top-relaxed} when each $\hat{F}_i$ is convex and Assumption \ref{bound_grad} holds. Suppose there exists a ball of radius $R_\mathcal{W}$ that contains $\mathcal{W}$. With step sizes $ \eta_w = {2R_\mathcal{W}}/({ (1+\alpha \hat{H}) \hat{L} \sqrt{nT}})$ and $\eta_p = {2}/(\hat{G}_p\sqrt{T})$, the output of TR-MAML satisfies:
\vspace{-0mm}
\begin{equation}
    \mathbb{E} \left[ \max_{p \in \Delta_n} \phi(w^\textbf{c}_T, p) -\min_{w \in \mathcal{W}} \phi(w,p^\textbf{c}_T) \right] \leq \frac{3\sqrt{n}R_\mathcal{W} (1+\alpha \hat{H}) \hat{L} + 3\hat{G}_p}{\sqrt{T}} \nonumber
\end{equation}
\end{theorem}
\vspace{0mm}
Thus, TR-MAML requires $T = \mathcal{O}(1/\epsilon^2)$ iterations to reach an expected duality gap of at most $\epsilon$. Since it computes a constant number of stochastic oracle evaluations per iteration, its convergence rate is the optimal $\mathcal{O}(1/\epsilon^2)$ stochastic oracle calls to reach an $\epsilon$-accurate solution. 

\noindent \textbf{Nonconvex Setting.} We next study the case when each $\hat{F}_i$ may be nonconvex and as a result, $\phi(w,p)$ may be nonconvex in $w$. Here we must evaluate the pair $(w^{\tau}_T,p^{\tau}_T)$ returned by our algorithm differently with respect to $p$ and $w$: we still intend that $p^\tau_T \in \Delta_n$ globally maximizes $\phi(w^{\tau}_T,\cdot)$, but can only hope to find $w^{\tau}_T$ near a stationary point of $\phi(\cdot,p^{\tau}_T)$. Thus, we say that $(\tilde{w}, \tilde{p})$ is an $(\epsilon, \delta)$-stationary point of $\phi$ if
\begin{equation}
\|\nabla_w \phi(\tilde{w},\tilde{p})\|_2  \leq \epsilon\quad \text{ and }\quad \phi(\tilde{w},\tilde{p}) \geq \max_{p \in \Delta_n} \phi(\tilde{w},p) - \delta,
\end{equation}
where $\epsilon, \delta > 0$, assuming that $\mathcal{W}=\mathbb{R}^d$, otherwise we consider the projected gradient, which we discuss later. In either case we will leverage smoothness. The function that we aim to minimize, $\max_{p\in \Delta_n} \phi(w,p)$, is non-smooth because of the maximization, but we can again adapt a result from \citep{finn2019online} to show that each $\hat{F}_i$ is smooth under the previous assumptions on each $\hat{f}_{i,j}$.
\begin{lemma} (Adapted from \citep{finn2019online}, Theorem 1) \label{lemma-smooth}
Under Assumptions \ref{bound_grad} and \ref{assump_hess}, each $\hat{F}_i$ is $\tilde{M}$-smooth, where $\tilde{M} \coloneqq  \hat{M}(1+\alpha \hat{M})^2 +  \alpha \hat{L} \hat{\rho} $.
\end{lemma}

We must also compute the expected squared deviation of the stochastic gradient $\hat{g}_w$, denoted by $\sigma_w^2$.
\begin{lemma} \label{lemma_var} For all $w\in \mathcal{W}$ and $p \in \Delta_n$, 
\vspace{0mm}
\begin{equation}
\sigma_w^2(w,p) \coloneqq \mathbb{E} [ \| \hat{g}_w(w,p) - g_w(w,p)  \|^2_2] = \frac{n}{C} \sigma^2(w,p) + \frac{n}{C} \sum_{i=1}^n \sigma_i^2(w,p) 
\vspace{0.5mm}
\end{equation}
where $\sigma^2(w,p) \coloneqq \sum_{i=1}^n \|p_i \nabla \hat{F}_i(w) - \frac{1}{n} \sum_{i'=1}^{n}p_{i'}\nabla \hat{F}_{i'}(w)\|_2^2$ and\\
$\sigma_i^2(w,p) \coloneqq \frac{p_i^2}{m_i} \sum_{j=1}^{m_i} \| (I - \alpha \nabla^2 \hat{f}_{i,j}(w, D^{\text{train}}_{i,j}))\nabla \hat{f}_{i,j}(w-\alpha \nabla^2 \hat{f}_{i,j}(w, D^{\text{train}}_{i,j}), D^{\text{test}}_{i,j})  - \nabla \hat{F}_i(w)  \|_2^2$.
\end{lemma}
\vspace{0mm}
Here $\sigma^2$ represents the inter-task variance and each $\sigma_i^2$ represents an intra-task variance. With $\sigma_w^2$ defined, the convergence of TR-MAML when $\mathcal{W} = \mathbb{R}^d$ can be shown via the following theorem.
\begin{theorem} \label{noncon-cor1}
If Assumptions \ref{bound_grad} and \ref{assump_hess} hold, $\mathcal{W} = \mathbb{R}^d$ and
$\eta_w^t = T^{-\beta}$, and $ \eta_p^t = \sqrt{2}(T^{2\beta}\hat{G}_p)^{-1}$
for all $t=1,\dots,T$ and any $\beta \in (0,\tfrac{1}{2})$,
and $T^\beta > {\tilde{M}}/{2}$,
then the output of Algorithm~\ref{alg1} satisfies
\begin{align} \label{noncon-lemma-1-eq}
    &\mathbb{E}\big[\|\nabla_w \phi(w_T^\tau, p_T^\tau)\|_2^2 \big] \leq \frac{ \phi(w^1,p^1) + \hat{B} + \sqrt{2n}\hat{B}  + 2\tilde{M}\sigma_w^2}{T^\beta-\tilde{M}/2},\nonumber \\
 &   \mathbb{E}\left[ \phi(w_T^\tau, p_T^\tau)  \right] \geq \max_{p \in \Delta_n} \left\{ \mathbb{E} \left[ \phi(w^\tau_T, p)  \right] \right\} - \hat{G}_p / ( \sqrt{2} T^{\min\{2\beta, 1-2\beta\}} ). \nonumber
\end{align}
\end{theorem}
\vspace{0mm}
Theorem \ref{noncon-cor1} shows that Algorithm~\ref{alg1} converges in expectation to an $(\epsilon, \delta)$-stationary point of $\phi$ in $\mathcal{O}(\max\{1/\epsilon^{2/\beta}, 1/\delta^{1/\min\{2\beta, 1-2\beta\}}\})$ stochastic gradient evaluations in the unconstrained setting.
Note that $\beta$ can be tuned to favor convergence with respect to $w$ or $p$. 
To treat convergence with respect to $w$ and $p$ equally, the optimal setting is $\beta = \tfrac{2}{5}$, yielding a convergence rate of $\mathcal{O}(\max\{1/\epsilon^{5}, 1/\delta^{5}\})$. 

We finally consider the case when $\mathcal{W}$ is a compact, convex set. In this setting 
the notion of an $(\epsilon, \delta)$-stationary point must be altered such that $\epsilon$ upper bounds the projected gradient, $\bar{g}_w$, defined as
\begin{equation}
    \bar{g}_w(w^t,p^t) \coloneqq \tfrac{1}{\eta_w^t}(w^t -  \Pi_\mathcal{W}(w^t- \eta_w^t \hat{g}_w(w^t,p^t)) ), \nonumber
\end{equation}
since this vector reveals how much the solution can be improved by moving within the feasible set.
In the following theorem, we choose $C$ as a function of $T$ to show convergence.
\begin{theorem} \label{noncon-thrm3}
Suppose Assumptions \ref{bound_grad} and \ref{assump_hess} hold. Let $\tilde{\sigma}_w^2 \coloneqq C\sigma_w^2$ and set $\eta_w^t = 1/{(2\tilde{M})}$ and $\eta_p^t = (T^{\beta}  \hat{B} \sqrt{n})^{-1}$
for $t \in [T]$, and the task batch size as $C = T^{\beta}$, for any $\beta \in (0, 1)$, then
\begin{align*}
  &  \mathbb{E} \left[ \|\bar{g}_w(w^\tau_T, p^\tau_T)\|_2^2 \right] \leq
    \frac{8\tilde{M}(\phi(w^{1}, p^{1}) + \hat{B})}{3T}+ \frac{8\tilde{M}\hat{B}\sqrt{n} + 4\tilde{\sigma}_w^2}{3T^{\beta}} , \\
   & \mathbb{E}\left[ \phi(w^\tau_T, p^\tau_T)  \right] \geq \max_{p \in \Delta_n} \left\{ \mathbb{E} \left[ \phi(w^\tau_T, p)  \right] \right\} - \frac{\hat{B}\sqrt{n}}{T^{\min\{\beta, 1-\beta\}}} .
\end{align*}
\end{theorem}
\vspace{0mm}
The number of stochastic gradient evaluations is now $\mathcal{O}(CT) = \mathcal{O}(T^{1+\beta})$, so Theorem \ref{noncon-thrm3} shows
Algorithm~\ref{alg1} converges to an $(\epsilon, \delta)$-stationary point after at most computation of $\mathcal{O}(\max\{ 1/\epsilon^{(2+2\beta)/\beta}, 1/\delta^{(1+\beta)/\min\{\beta, 1-\beta\}} \})$ stochastic gradients with convex, compact $\mathcal{W}$ and nonconvex $\hat{F}_i$. By setting $\beta = \tfrac{2}{3}$ we treat convergence with respect to $w$ and $p$ equally, yielding a complexity of $\mathcal{O}(\max\{ 1/\epsilon^{5}, 1/\delta^{5} \})$ evaluations.
\section{Generalization Bounds} \label{sec:generalization}
Given that the meta-learner has access to a finite number of task instances during meta-training, there are two types of generalization to consider: generalization to new instances of previously-seen tasks, and generalization to new tasks. We start by bounding the error on new instances of previously-seen tasks. 
Note that each task's $\mathcal{D}_i$ is a distribution over $\mathcal{Z} \coloneqq (\mathcal{X} \times \mathcal{Y})^{K+J} $. 
For some loss $\ell$, define the family of functions $\mathcal{F}(\mathcal{Z}) \coloneqq \mathcal{F} \coloneqq \{ \hat{f}(w - \alpha \nabla \hat{f}(w, D^{\text{train}}), D^{\text{test}}) : w \in \mathcal{W} \}$, where $(D^{\text{train}}, D^{\text{test}}) \in \mathcal{Z}$ and $\hat{f}(w,D)$ is the average loss of $w$ on the samples in $D$. The Rademacher complexity of $\mathcal{F}$ on $m_i$ samples $\{ (D^{\text{train}}_j, D^{\text{test}}_j )\}_{j=1}^{m_i} =: \mathbf{D}$ drawn i.i.d. from $\mathcal{D}_i$ is then
\begin{equation}
\mathfrak{R}_{m_i}^i(\mathcal{F}) = \mathbb{E}_{\mathbf{D} \sim (\mathcal{D}_i )^{m_i} } \mathbb{E}_{\epsilon_j} \bigg[ \sup_{w \in \mathcal{W}} \frac{1}{m_i} \sum_{j=1}^{m_i} \epsilon_j \hat{f}_{i,j}(w - \alpha \nabla \hat{f}_{i,j}(w, D^{\text{train}}_{i,j}), D^{\text{test}}_{i,j}) \bigg],
\end{equation}
where the $\epsilon_j$'s are Rademacher random variables. Recall that the empirical loss of the model $w$ on the $i$-th task is
$\hat{F}_i(w)$, defined in \eqref{top-maxi_emp}.
By a standard Rademacher complexity bound, one can bound the analogous expected loss $F_i(w)$ with high probability over the choice of task instances.
\begin{proposition} \label{gen1}
Suppose Assumption \ref{bound_grad} holds, then with probability at least $1-\delta$,
\begin{equation}
{F}_i(w) \coloneqq \mathbb{E}_{(D_{i,j}^{\text{train}}, D_{i,j}^{\text{test}}) \sim \mathcal{D}_i}[ \hat{f}_{i,j} (w- \alpha \nabla \hat{f}_{i,j}(w, D_{i,j}^\text{train}), D_{i,j}^\text{test})] \leq \hat{F}_i(w) + 2\mathfrak{R}^i_{m_i}(\mathcal{F}) + \hat{B} \sqrt{\frac{\log 1/\delta}{2m_i} } \nonumber
\end{equation}
\end{proposition}
Next, let $w^\ast$ be the optimal solution to the TR-MAML meta-training objective \eqref{top-relaxed}. 
Suppose a new task is drawn with distribution $\mathcal{D}_{n+1}$, and suppose that $\mathcal{D}_{n+1} = \sum_{i=1}^{n} a_i \mathcal{D}_{i}$ for some $a \in \Delta_n$. Then the loss $F_{n+1}(w)$ is a convex combination of the losses on the meta-training tasks, yielding
\begin{theorem} \label{gen_bound}
For a new task with distribution $\mathcal{D}_{n+1}$, if $\mathcal{D}_{n+1} = \sum_{i=1}^{n} a_i \mathcal{D}_{i}$ for $a \in \Delta_n$, then
\begin{equation}
{F}_{n+1}(w^\ast)  \leq \min_{w \in \mathcal{W}} \max_{p \in \Delta_n} \sum_{i=1}^n p_i \hat{F}_i(w)  + 2a_i\mathfrak{R}^i_{m_i}(\mathcal{F}) + a_i \hat{B} \sqrt{\frac{\log (n/\delta)}{2m_i} }
\end{equation}
\end{theorem}
Theorem \ref{gen_bound} shows that the min-max meta-training solution leverages the diversity of the meta-training tasks to generalize across their full convex hull, not just a local neighborhood of the solution.

\vspace{-1mm}
\section{Experimental Results} \label{section:experiments}
\vspace{-1mm}
Our experiments study whether minimizing the maximum loss during meta-training leads to a more task-robust solution compared to MAML in few-shot sinusoid regression and image classification.

\begin{figure}[t]
\begin{center}
\includegraphics[width=1\columnwidth]{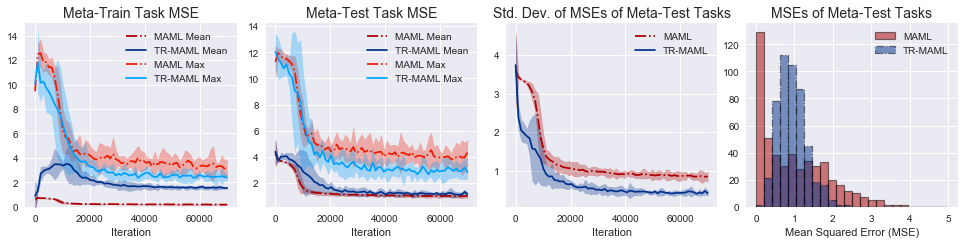}
\caption{Meta-training and meta-test task MSE statistics vs the number of meta-training iterations for $K=5$, with 95\% confidence intervals shaded over 5 trials. The rightmost plot shows the number of meta-test tasks with average MSE within particular intervals for a sample trial. TR-MAML outperforms MAML on the worst-case regression task, and performs more uniformly across all tasks.}
\label{fig:sine}
\end{center}
\end{figure}

\begin{table}
  \caption{Sinusoid regression results showing MSE statistics across the 490 meta-test tasks for the model updated by one step of SGD using $K \in \{5, 10\}$ samples from the initialization resulting from 70,000 meta-training iterations. To estimate the 490 meta-test task losses, 5,000 total task instances were evaluated, and 95\% confidence intervals are given over 3 meta-training and meta-testing runs.}
  \vspace{3mm}
  \label{sine-table}
  \centering
  \begin{tabular}{lclclclc}
    \toprule
         &  Algorithm & Mean & Worst & Std. Dev. \\
    \midrule
        \multirow{2}{*}{5-shot}  & MAML  & $\mathbf{1.02\pm 0.10}$ & $3.89\pm 0.83$ & $0.88\pm 0.14$    \\
             & TR-MAML & $1.09\pm 0.08 $ & $\mathbf{2.82 \pm 0.35}$ &  $\mathbf{0.43\pm 0.03}$       \\   
     \cmidrule(r){1-5}
     \multirow{2}{*}{10-shot}  & MAML  & $ \mathbf{0.66 \pm 0.16}$ & $ 2.57 \pm 0.70$  & $ 0.54 \pm 0.13$    \\
             & TR-MAML & $0.77 \pm 0.11$ & $\mathbf{1.68 \pm 0.43 }$ &  $\mathbf{0.25 \pm 0.08 }$     \\   
      \bottomrule
  \end{tabular}
\end{table}
\subsection{Sinusoid Regression} In the popular sinusoid regression experiment \citep{finn2017model}, each task instance is a sinusoid regression problem in which the target is a sine function on $[-5, 5] \subset \mathbb{R}$ with amplitude $a \in [0.1,5]$ and phase $b \in [0,2\pi]$. 
The learner has $K$ samples $\{(x_i, a\sin(x_i-b))\}_{i=1}^K$, where each $x_i$ is uniformly sampled from $[-5,5]$, and tries to find a function 
that closely approximates $a \sin(x-b)$ in terms of mean squared error (MSE). Typically the meta-training and meta-testing distributions are identical, and are such that amplitudes are drawn uniformly from $[0.1, 5]$ and phases uniformly from $[0,2\pi]$. Here we experiment with a distributional shift between meta-training and meta-testing in which a large number of easy task instances and a small number of hard task instances are accessible for meta-training, and the resulting initialization is evaluated on all types of tasks in the space, a common learning scenario.
In particular, we assume that sine functions of all phases but with amplitudes only in the intervals $[0.1,1.05]$ (easy tasks) and $[4.95,5]$ (hard tasks) are available for meta-training. The sinusoids with larger amplitudes are harder targets because they are less smooth and have larger magnitudes, meaning poor approximations are generally punished more severely in terms of MSE. Empirically we find that phase has little effect on the hardness of a target.
To implement TR-MAML, we partition $[0.1,5]$ into 490 disjoint subintervals of length 0.01, and define a task to be the uniform distribution over all task instances with target amplitude in a particular subinterval. Thus there are 95 easy and 5 hard meta-training tasks. We assume each task has the same number of instances available, so both MAML and TR-MAML sample phases uniformly from $[0,2\pi]$ and amplitudes uniformly from $[0.1,1.05] \cup [4.95,5]$. The meta-test distribution is the uniform distribution across the full space of amplitudes and phases. Both algorithms use one SGD step as the inner learning algorithm, and a fully-connected neural network with two hidden layers composed of 40 nodes with ReLU activations for the learning model, equivalent to the one in \citep{finn2017model}.

Figure \ref{fig:sine} shows the convergence trajectories of MAML and TR-MAML when $K=5$. Each plot entails estimating the current model's MSE on each task by sampling 5,000 task instances across all 100 meta-training tasks (for an average of 50 instances per task), and separately across all 490 meta-testing tasks. 
The leftmost plot shows the average and maximum MSE over each of the 100 meta-training tasks' estimated MSE vs the number of iterations, and the middle-left plot shows the same statistics over the 490 meta-testing tasks. During meta-training, TR-MAML sacrifices average for worst-case task performance. However, its focus on task-robustness yields more uniform performance across all tasks, allowing TR-MAML to outperform MAML on the hardest meta-test tasks while nearly matching MAML's average performance after the distributional shift. TR-MAML's more uniform performance for $K=5$ is captured in the middle-right plot of Figure \ref{fig:sine}, which shows the standard deviation across the meta-testing task MSEs vs the number of iterations, and the rightmost plot of Figure \ref{fig:sine}, a histogram of the average MSEs among the 490 meta-test tasks, computed using 5,000 total task instances.
Table \ref{sine-table} tells a similar story for the $K\in \{5,10\}$-shot cases by giving the average, maximum, and standard deviation of the MSEs among the 490 meta-test tasks after full meta-training, where the statistics are again empirical averages over 5,000 task instances. 

\begin{table}
  \caption{Omniglot $N$-way, $K$-shot classification accuracies (\%). After meta-training, 5,000 few-shot classification problems (task instances) are sampled uniformly from the 25 alphabets (tasks) used for meta-training, likewise for the 20 new meta-testing alphabets. For each alphabet, the average accuracy on task instances from that alphabet is computed, and statistics are taken across these average accuracies. `Weighted Mean' weighs the alphabet accuracies by the meta-training distribution, which corresponds to the quantity MAML aims to optimize, whereas `Mean' weighs all alphabets equally. `Worst' is the minimum alphabet accuracy, and `Std. Dev.' is the standard deviation across the alphabet accuracies, with 95\% confidence intervals given over three full runs for all statistics.}
   \vspace{1mm}
  \label{omni-table}
  \centering
  \resizebox{\columnwidth}{!}{%
  \begin{tabular}{llllllll}
    \toprule
   \multicolumn{2}{c}{}  & \multicolumn{3}{c}{Meta-training Alphabets}  &  \multicolumn{3}{c}{Meta-testing Alphabets}               \\
    \cmidrule(r){3-5}
    \cmidrule(r){6-8}
      $(N,K)$   &  Algorithm & Weighted Mean & Mean & Worst & Mean & Worst & Std. Dev. \\
    \midrule
        \multirow{2}{*}{(5,1)}  & MAML  & $\mathbf{98.4\pm.2}$  &$96.6\pm.2$ & $82.4\pm 1.1$    & $\mathbf{93.5\pm .2}$  & $82.5\pm .2$   & $3.84\pm .1$   \\
             & TR-MAML &  $97.4\pm.6$  &  $\mathbf{97.5\pm.1}$ & $\mathbf{95.0\pm0.3}$   & $93.1\pm1.1$  & $\mathbf{85.3\pm1.9}$  & $\mathbf{3.50\pm .3}$    \\   
\cmidrule(r){1-8}
     \multirow{2}{*}{(20,1)}  & MAML  & $\mathbf{99.2\pm .1}$ &$76.5 \pm .8$ & $33.9 \pm 3.0$ &  $67.6 \pm 2.0$ & $49.7 \pm 3.5$ &  $9.10 \pm .1$    \\
             & TR-MAML & $92.2\pm .8$  & $\mathbf{90.0\pm .9}$  & $\mathbf{82.4 \pm 2.1}$ & $\mathbf{74.3 \pm 1.4}$ & $\mathbf{58.4 \pm 1.8}$ & $\mathbf{8.70 \pm .5}$    \\   
      \cmidrule(r){1-8}
     \multirow{2}{*}{(20,5)}  & MAML  & $\mathbf{99.7 \pm .1}$ & $86.4 \pm .2$  & $50.3 \pm 1.3$ & $80.0 \pm .7$ & $67.9 \pm 1.9$ & $6.84 \pm .5$   \\
             & TR-MAML & $98.5 \pm .2$& $\mathbf{98.0 \pm .1}$  & $\mathbf{94.6 \pm .5}$ & $\mathbf{87.6 \pm .6}$ & $\mathbf{78.2 \pm .3}$ &  $\mathbf{5.39 \pm .3}$   \\   
    \bottomrule
  \end{tabular}%
  }
\end{table}

\subsection{Image Classification}
In few-shot image classification, the task instances are $N$-way, $K$-shot classification problems, where $N$ is the number of classes and $K$ is the number of labelled samples from each class that are available to the learner. After updating the model based on these $NK$ samples, the model is evaluated on $J$ samples from each class. We experiment in this setting using the Omniglot dataset \citep{lake2015human}. Omniglot contains 1623 handwritten characters from 50 alphabets, with 20 examples per character. As noted by \citet{triantafillou2019meta}, few-shot classification of Omniglot characters sampled from different alphabets has become too easy for modern techniques. Following their example, we instead consider more difficult, fine-grained task instances in which each class is a character from the \textit{same} alphabet. Each task is defined as the uniform distribution over all task instances composed of characters from a particular alphabet. We use the same splits as \citet{triantafillou2019meta}, which are the original splits proposed by \citet{lake2015human} with the 5 smallest alphabets used for meta-validation. There are $n=25$ alphabets, i.e., tasks, for meta-training and 20 for meta-testing. Suppose there are $Z_i$ characters in the $i$-th alphabet, then the number of task instances that may be drawn from the $i$-th task is proportional to ${Z_i \choose N}$, since every character has the same number of samples (20). These proportions define the empirical distribution over the $25$ meta-training tasks, so during meta-training MAML samples task instances by first selecting the $i$-th alphabet with probability proportional to ${Z_i \choose N}$, then uniformly samples an $N$-way, $K$-shot classification problem from the available data in alphabet $i$. Conversely, TR-MAML first samples an alphabet uniformly, then samples an $N$-way, $K$-shot problem uniformly from that alphabet. 

We use the same 4-layer CNN used in the original MAML paper \citep{finn2017model}. After 60,000 meta-training iterations, we evaluate the models yielded by MAML and TR-MAML on 5,000 $N$-way, $K$-shot classification problems from the 20 meta-test alphabets, as well as 5,000 problems from the meta-training alphabets.
Table \ref{omni-table} shows statistics taken over the average accuracy on task instances from each alphabet for different combinations of $N$ and $K$. 
Note that the empirical distribution of meta-training alphabets becomes more skewed as $N$ increases, causing MAML to focus on a smaller subset of the meta-training alphabets and further disregard worst-case alphabet performance. In contrast, TR-MAML prioritizes performance on all of the meta-training alphabets equally, leading to an `alphabet-robust' model that typically outperforms MAML on new alphabets, not only in the worst case but also on average across alphabets. Additional experimental results are given in Appendix \ref{app:experiments}.

  

\section{Conclusion}
Worst-case task performance is critical in many real-world learning systems, yet the meta-learning literature has not, up until now, produced a meta-learning procedure that optimizes for performance on the worst-case task. In this paper, we presented a novel variant of the MAML formulation for learning how to learn optimally on the worst-case task from some environment. We gave an algorithm to solve this formulation and proved meta-training convergence results in both convex and nonconvex settings, as well as a generalization result bounding the error on new tasks at meta-test time. Moreover, our experimental results demonstrated our algorithm's significant improvements over MAML in terms of robustness to the worst-case task and to shifts in the distribution over tasks between meta-training and meta-testing. We hope that our work is a starting point for future studies improving on the worst-case performance and distributional robustness of meta-learning systems.

\clearpage
\appendix

\begin{center}
\LARGE{\textbf{Appendix}}
\end{center}

\vspace{5mm}

\section{Formal Statement of Assumptions}

\textbf{Assumption 1} For all $i \in [n]$ and $j \in [m_i]$, $\hat{f}_{i,j}(\cdot, D^{\text{test}}_{i,j})$ satisfy the following:
\begin{enumerate}[topsep=1pt,itemsep=0ex,partopsep=1ex,parsep=1ex]
	\item $\hat{B}$-boundedness: $\exists \hat{B} \in \mathbb{R}$ s.t. $\forall w \in \mathcal{W}$,
 $|\hat{f}_{i,j}(w, D^{\text{test}}_{i,j})| \leq \hat{B}$.
    \item $\hat{L}$-Lipschitz continuity: $\exists \hat{L} \in \mathbb{R}$ s.t. $\forall u,v \in \mathcal{W}$,
 $|\hat{f}_{i,j}(u, D^{\text{test}}_{i,j}) - \hat{f}_{i,j}(v, D^{\text{test}}_{i,j}) | \leq \hat{L}\|u-v\|_2$.
\end{enumerate}
Furthermore, each $\hat{f}_{i,j}(\cdot, D^{\text{train}}_{i,j})$ satisfies the following:
\begin{enumerate}
 \item Hessian eigenvalue lower bound: $\exists \hat{H} \in \mathbb{R}$ s.t. $ \forall w \in \mathcal{W}$, $\lambda_{\min}(\nabla^2 \hat{f}_{i,j}(w,D^{\text{train}}_{i,j})) \geq -\hat{H}$.
\end{enumerate}

\noindent \textbf{Assumption 2} For all $i \in [n]$ and $j \in [m_i]$, $\hat{f_i}(\cdot, D^{\text{train}}_{i,j})$ satisfies the following:
 \begin{enumerate}
  \item $\hat{M}$-smoothness: $\exists \hat{M} \in \mathbb{R}$ s.t. $\forall u,v \in \mathcal{W}$, $\|\nabla \hat{f}_{i,j}(u, D^{\text{train}}_{i,j}) - \nabla \hat{f}_{i,j}(v,D^{\text{train}}_{i,j})\|_2 \leq \hat{M}\|u-v\|_2$.
 \item $\hat{\rho}$-Hessian-Lipschitz continuity: $\exists \hat{\rho} \in \mathbb{R}$ s.t. $\forall u,v \in \mathcal{W}$,
 $|\nabla^2 \hat{f}_{i,j}(u, D^{\text{train}}_{i,j}) - \nabla^2 \hat{f}_{i,j}(v, D^{\text{train}}_{i,j}) | \leq \hat{\rho}\|u-v\|_2$.
 \end{enumerate}
 
\section{Proof of Lemma \ref{lemma_unbiased}}

\subsection{Unbiasedness}

\begin{proof}
Recall that $\hat{g}_w(w,p)$ is computed as follows:
\begin{align} 
\hat{g}_w(w,p) &= \frac{n}{C} \sum_{k=1}^C p_{i_k}  (I - \alpha \nabla^2 \hat{f}_{i_k,j_k}(w, D_{i_k,j_k}^{\text{train}})) \nabla \hat{f}_{i_k,j_k}(w - \alpha \nabla \hat{f}_{i_k,j_k}(w, D_{i_k,j_k}^{\text{train}}), D_{i,j_i}^{\text{test}}) 
\nonumber
\end{align}
Thus we have
\begin{align}
&\mathbb{E}[\hat{g}_{w}(w,p)] \nonumber\\
&= \mathbb{E}_{\{ (i_k, j_k) \}_k }\left[\frac{n}{C} \sum_{k=1}^C p_{i_k}  (I - \alpha \nabla^2 \hat{f}_{i_k, j_k}(w, D_{i_k,j_k}^{\text{train}})) \nabla \hat{f}_{i_k, j_k}(w - \alpha \nabla \hat{f}_{i_k,j_k}(w, D_{i_k,j_k}^{\text{train}}), D_{i_k,j_k}^{\text{test}}) \right] \nonumber \\
&\stackrel{a}{=}  \mathbb{E}_{\{ i_k \}_{k}} \left[ \frac{n}{C} \sum_{k=1}^C  \mathbb{E}_{\{ j_k\}_{k}} \left[p_{i_k}  (I - \alpha \nabla^2 \hat{f}_{i_k,j_k}(w, D_{i_k,j_k}^{\text{train}})) \nabla \hat{f}_{i_k,j_k}(w - \alpha \nabla \hat{f}_{i_k,j_k}(w, D_{i_k, j_k}^{\text{train}}), D_{i_k,j_k}^{\text{test}}) \mid {\{ i_k \}}_{k} \right] \right]  \nonumber \\
&= \mathbb{E}_{\{ i_k \}_{k}} \left[\frac{n}{C} \sum_{k=1}^C  \frac{p_{i_k}}{m_{i_k}} \sum_{j=1}^{m_{i_k}}  (I - \alpha \nabla^2 \hat{f}_{i_k,j}(w, D_{i_k,j}^{\text{train}})) \nabla \hat{f}_{i_k,j}(w - \alpha \nabla \hat{f}_{i_k,j}(w, D_{i_k,j}^{\text{train}}), D_{i_k,j}^{\text{test}}) \mid \mathcal{C} \right] \nonumber \\
&\stackrel{b}{=}  \sum_{i=1}^n \frac{p_{i}}{m_i} \sum_{j=1}^{m_i} (I - \alpha \nabla^2 \hat{f}_{i,j}(w, D_{i,j}^{\text{train}})) \nabla \hat{f}_{i,j}(w - \alpha \nabla \hat{f}_{i,j}(w, D_{i,j}^{\text{train}}), D_{i,j}^{\text{test}})  \nonumber \\
&= g_w(w,p)
\end{align}
where $a$ follows from the Law of iterated Expectation and $b$ follows because each index $i_k$ is selected with probability $1/n$. A similar computation shows that $\mathbb{E}[\hat{g}_p(w,p)] = g_p(w,p)$.
\end{proof}

\subsection{Bounded Second Moments} \label{app:bound_grad}

\begin{proof}
First we show the bound on $\mathbb{E}[\| \hat{g}_p(w,p) \|_2^2]$.
Recall that $\hat{g}_p(w,p) = \sum_{k=1}^C  \frac{n}{C}\hat{f}_{i_k} (w -  \alpha \nabla \hat{f}_{i_k}(w,D^{\text{train}}_{i_k,j_k}), D^{\text{test}}_{i_k,j_k})e_{i_k}$. Let $c_i$ be the number of times index $i$ appears in $\{i_k\}_{k=1}^C$.

Then, noting that each $c_i$ is a binomial random variable with success probability $\frac{1}{n}$ and $C$ trials, we have
\begin{align}
\mathbb{E}[\| \hat{g}_p(w,p) \|_2^2] &= \sum_{i=1}^n \mathbb{E}[(\hat{g}_p(w,p)_i)^2] \nonumber \\
&= \sum_{i=1}^n \mathbb{E}[(\sum_{k\in [C] : i_k = i}  \frac{n}{C}\hat{f}_{i} (w -  \alpha \nabla \hat{f}_{i}(w,D^{\text{train}}_{i,j_k}), D^{\text{test}}_{i,j_k}))^2] \nonumber \\
&\leq   \frac{n^2}{C^2} \sum_{i=1}^n \mathbb{E}[(\sum_{k\in [C] : i_k = i} \hat{B})^2]  \label{eqcs}\\
&= \frac{n^2}{C^2} \sum_{i=1}^n \mathbb{E}[(c_i \hat{B})^2] \nonumber \\
&= \frac{n^2}{C^2} \sum_{i=1}^n (\frac{C(n-1)}{n^2} + \frac{C^2}{n^2} )\hat{B}^2 \nonumber \\
&= \frac{n}{C} (n+ C-1)\hat{B}^2 =: \hat{G}_p^2 \label{gpdef}
\end{align}
where \eqref{eqcs} follows from Assumption \ref{bound_grad}. Next we bound $\mathbb{E}[\| \hat{g}_w(w,p) \|_2^2]$.
Recall the definition of $\hat{g}_w(w,p)$:
\begin{equation}
\hat{g}_w(w,p) =   \sum_{k =1}^C \frac{n}{C} p_{i_k} (I - \alpha \nabla^2 \hat{f}_{i_k,j_k}(w, D_{i_k,j_k}^{\text{train}})) \nabla \hat{f}_{i_k,j_k}(w - \alpha \nabla \hat{f}_{i_k,j_k}(w, D_{i_k,j_k}^{\text{train}}), D_{i_k,j_k}^{\text{test}})
\end{equation}
We can write $\hat{g}_w(w,p)= \frac{n}{C}\sum_{k=1}^C X_k$ where $X_k$ is written as
\begin{equation}
X_k =p_{i_k} (I - \alpha \nabla^2 \hat{f}_{i_k,j_k}(w, D_{i_k,j_k}^{\text{train}})) \nabla \hat{f}_{i_k,j_k}(w - \alpha \nabla \hat{f}_{i_k, j_k}(w, D_{i_k,j_k}^{\text{train}}), D_{i_k,j_k}^{\text{test}}) \label{defX}
\end{equation}
We have
\begin{align}
\mathbb{E} [\| X_k\|_2^2 ] &=  \frac{1}{n} \sum_{i=1}^n \frac{1}{m_i} \sum_{j=1}^{m_i}  p_{i}^2 \|  (I - \alpha \nabla^2 \hat{f}_{i,j}(w, D_{i,j}^{\text{train}})) \nabla \hat{f}_{i,j}(w - \alpha \nabla \hat{f}_{i,j}(w, D_{i,j}^{\text{train}}), D_{i,j}^{\text{test}})   \|_2^2 \nonumber \\
&\leq \frac{1}{n}\sum_{i=1}^n \frac{p_i^2}{m_i} \sum_{j=1}^{m_i}  \|  (I - \alpha \nabla^2 \hat{f}_{i,j}(w, D_{i,j}^{\text{train}})) \|_2^2 \| \nabla \hat{f}_{i,j}(w - \alpha \nabla \hat{f}_{i,j}(w, D_{i,j}^{\text{train}}), D_{i,j}^{\text{test}})   \|_2^2 \label{var1} \\
&\leq \frac{1}{n} \sum_{i=1}^n \frac{p_i^2}{m_i} \sum_{j=1}^{m_i}  (1+\alpha \hat{H})^2  \| \nabla \hat{f}_{i,j}(w - \alpha \nabla \hat{f}_{i}(w, D_{i,j}^{\text{train}}), D_{i,j}^{\text{test}})   \|_2^2  \label{var2}\\
&\leq \frac{1}{n}\sum_{i=1}^n \frac{p_i^2}{m_i} \sum_{j=1}^{m_i}  (1+\alpha \hat{H})^2 \hat{L}^2 \label{var3} \\
&\leq \frac{1}{n}(1+\alpha \hat{H})^2 \hat{L}^2 \label{var4}
\end{align}
where \eqref{var1} follows by the Cauchy-Schwarz Inequality, \eqref{var2} and \eqref{var3}  follow from Assumption \ref{bound_grad}, and \eqref{var4} follows from the fact that $\sum_{i=1}^n p_i^2 \leq 1$. Thus we have
\begin{align}
\mathbb{E}[\| \hat{g}_{w}(w,p) \|_2^2 ] &= \mathbb{E} [  \| \frac{n}{C} \sum_{k=1}^C X_k\|_2^2 ] \\
&\leq  \mathbb{E} [ \frac{n^2}{C}\sum_{k = 1}^C  \| X_k\|_2^2 ] \label{var5} \\
&\leq {n} (1+\alpha \hat{H})^2 \hat{L}^2
\end{align}
where \eqref{var5} follows from the convexity of norms and Jensen's Inequality. 
\end{proof}

\section{Proof of Theorem \ref{theorem_con}}

\begin{proof}
We adapt the arguments from \citep{mohri2019agnostic} to our nested gradients case. First observe that since each $\hat{F}_i(w)$ is convex, $\phi(w,p)$ is convex in $w$ and linear, thus concave, in $p$. Therefore we can write:
\begin{align}
    \max_{p \in \Delta_n} \phi(w^\mathbf{c}_T, p) - \min_{w \in \mathcal{W} }\phi(w, p^\mathbf{c}_T) &= \max_{p \in \Delta_n} \left\{ \phi(w^\mathbf{c}_T, p) - \min_{w \in \mathcal{W} }\phi(w, p^\mathbf{c}_T)  \right\} \nonumber \\
    &= \max_{p \in \Delta, w \in \mathcal{W}} \left\{ \phi(w^\mathbf{c}_T, p) - \phi(w, p^\mathbf{c}_T)  \right\} \nonumber \\
    &\leq \dfrac{1}{T} \max_{p \in \Delta, w \in \mathcal{W}} \left\{ \sum_{t=1}^T \phi(w^t, p) - \phi(w, p^t)  \right\} \label{convexity}
\end{align}
where \eqref{convexity} follows from the convexity of $\phi$ in $w$ and the concavity of $\phi$ in $p$. Again using the convexity of $\phi$ in $w$ along with the linearity of $\phi$ in $p$, we have that for any $t \geq 1$,
\begin{align}
    \phi(w^t, p) - \phi(w, p^t) &= \phi(w^t,p) - \phi(w^t,p^t) + \phi(w^t,p^t) - \phi(w, p^t) \nonumber \\
    &\leq \langle (p-p^t), \nabla_p \phi(w^t,p^t) \rangle + \langle (w^t-w), \nabla_w \phi(w^t,p^t)\rangle \nonumber \\
    &= \langle (p-p^t), \hat{g}_p^t \rangle + \langle(w^t-w), \hat{g}_w^t \rangle \nonumber \\
    &\quad + \langle (p-p^t), (\nabla_p \phi(w^t,p^t) - \hat{g}_p^t) \rangle + \langle(w^t-w),  (\nabla_w \phi(w^t,p^t) - \hat{g}_w^t)\rangle \nonumber
\end{align}
Thus by rearranging terms and the subadditivity of $\max$,
\begin{align}
   \max_{p \in \Delta, w \in \mathcal{W}} &\left\{ \sum_{t=1}^T \phi(w^t, p) - \phi(w, p^t) \right\} \nonumber\\
  &\leq \max_{p \in \Delta, w \in \mathcal{W}} \left\{ \sum_{t=1}^T \langle (p-p^t), \hat{g}_p^t\rangle + \langle(w^t-w), \hat{g}_w^t \rangle \right\} \nonumber \\
    &\quad + \max_{p \in \Delta, w \in \mathcal{W}} \left\{ \sum_{t=1}^T \langle p, (\nabla_p \phi(w^t,p^t) - \hat{g}_p^t)\rangle + \langle w, (\hat{g}_w^t- \nabla_w \phi(w^t,p^t) )\rangle \right\} \nonumber \\
    &\quad - \left(\sum_{t=1}^T \langle p^t, (\nabla_p \phi(w^t,p^t) - \hat{g}_p^t)\rangle - \langle w^t, (\nabla_w \phi(w^t,p^t) - \hat{g}_w^t) \rangle \right) \label{terms}
\end{align}
We bound the expectation of each of the above terms separately, starting with the first one. Note that since $2ab = a^2 + b^2 - (a-b)^2$, we have that for any $w \in \mathcal{W}$ and constant step size $\eta_w > 0$,
\begin{align}
    \sum_{t=1}^T \langle (w^t - w),\hat{g}_w^t\rangle &= \dfrac{1}{2} \sum_{t=1}^T \dfrac{1}{\eta_w}\| w^t - w\|_2^2 + \eta_w \|\hat{g}_w^t\|_2^2 - \dfrac{1}{\eta_w}\|w^t - \eta_w \hat{g}_w^t - w\|_2^2 \nonumber \\
    &\leq \dfrac{1}{2\eta_w} \sum_{t=1}^T \| w^t - w\|_2^2 + (\eta_w)^2 \|\hat{g}_w^t\|_2^2 - \|w^{t+1} - w\|_2^2 \label{proj_prop2} \\
    &= \dfrac{1}{2\eta_w} (\| w^1 - w\|_2^2 - \|w^{T+1} - w\|_2^2) + \dfrac{\eta_w}{2} \sum_{t=1}^T\|\hat{g}_w^t\|_2^2 \label{telescop} \\
    &\leq \dfrac{1}{2\eta_w} \| w^1 - w\|_2^2 + \dfrac{\eta_w}{2}\sum_{t=1}^T \|\hat{g}_w^t\|_2^2 \nonumber \\
    &\leq \dfrac{2 R_\mathcal{W}^2}{\eta_w} + \dfrac{\eta_w}{2}\sum_{t=1}^T \|\hat{g}_w^t\|_2^2 \label{last}
\end{align}
where \eqref{proj_prop2} follows from the projection property and \eqref{telescop} is the result of the telescoping sum. Since \eqref{last} holds for all $w\in \mathcal{W}$ and its right hand side does does not depend on $w$, we can take the maximum over $w \in \mathcal{W}$ on the left hand side, and the expectation of both sides with respect to the stochastic gradients, to obtain
\begin{equation}
    \mathbb{E}\left[ \max_{w \in \mathcal{W}} \sum_{t=1}^T \langle (w^t - w), \hat{g}^t_w \rangle  \right] \leq \dfrac{2 R_\mathcal{W}^2}{\eta_w} + \dfrac{\eta_w T\hat{G}_w^2}{2}
\end{equation}
where $\hat{G}_w^2 = n (1+ \alpha \hat{H})^2\hat{L}^2$ is the bound on the second moment of the stochastic gradient with respect to $w$ given in Lemma \ref{lemma_unbiased}.
Using analogous arguments and noting that the radius of $\Delta_n$ is 1, we can show that 
\begin{equation}
    \mathbb{E}\left[ \max_{p \in \Delta_n} \sum_{t=1}^T \langle (p - p^t), \hat{g}^t_p \rangle  \right] \leq \dfrac{2}{\eta_p} + \dfrac{\eta_p T\hat{G}_p^2}{2}
\end{equation}
where, again from Lemma \ref{lemma_unbiased}, $\hat{G}_p^2 = \frac{n(n+C-1) \hat{B}^2}{C}$.
Next, for the second term in \eqref{terms}, we can use the Cauchy-Schwarz Inequality and again the fact that $\max_{p \in \Delta_n} \|p\|_2 =1$ to write
\begin{align}
    \max_{p \in \Delta_n} \sum_{t=1}^T \langle p, \nabla_p \phi(w^t, p^t) - \hat{g}_p^t \rangle &= \max_{p \in \Delta_n} \langle p, \sum_{t=1}^T \nabla_p \phi(w^t, p^t) - \hat{g}_p^t \rangle \nonumber \\
    &\leq \|\sum_{t=1}^T  \nabla_p \phi(w^t, p^t) - \hat{g}_p^t  \|_2 \label{afl}
\end{align}
Note from Lemma \ref{lemma_unbiased} that
\begin{align} \label{def_sigmap}
\mathbb{E}[\| \nabla_p \phi(w^t, p^t) - \hat{g}_p^t  \|_2^2] &= \mathbb{E}[\| \hat{g}_p^t  \|_2^2] - \| \nabla_p \phi(w^t, p^t)\|^2_2 \nonumber \\
&\leq \hat{G}_p^2 \nonumber
\end{align}
for all $t \geq 1$.
Define $\tilde{\sigma}_p^2$ such that $\mathbb{E}[\| \nabla_p \phi(w^t, p^t) - \hat{g}_p^t  \|_2^2]  \leq \tilde{\sigma}_p^2 \leq \hat{G}_p^2$ for all $t \geq 1$.
Also note that because the batch selections are independent, the $\nabla_p \phi(w^t, p^t) - \hat{g}_p^t$ terms are uncorrelated random variables with mean 0.
Using this fact combined with the definition of $\tilde{\sigma}_p^2$, we obtain
\begin{align*}
    \mathbb{E}[\| \sum_{t=1}^T  \nabla_p \phi(w^t, p^t) - \hat{g}_p^t  \|_2]^2 &\leq \mathbb{E}[\|\sum_{t=1}^T  \nabla_p \phi(w^t, p^t) - \hat{g}_p^t  \|_2^2] \\
    &= \mathbb{E}[ \sum_{t=1}^T \|  \nabla_p \phi(w^t, p^t) - \hat{g}_p^t  \|_2^2] \\
    &\leq T\tilde{\sigma}_p^2 
\end{align*}
which implies that $\mathbb{E}[\|\sum_{t=1}^T  \nabla_p \phi(w^t, p^t) - \hat{g}_p^t  \|_2] \leq \sqrt{T}\tilde{\sigma}_{p}$. Using this relation after taking the expectation of both sides of \eqref{afl} yields
\begin{equation}
    \mathbb{E} \left[ \max_{p \in \Delta_n} \sum_{t=1}^T \langle p, \nabla_p \phi(w^t, p^t) - \hat{g}_p^t \rangle \right] \leq \sqrt{T}\tilde{\sigma}_p
\end{equation}
Using similar arguments and the analogous definition of $\tilde{\sigma}_w^2$, with this time using $R_\mathcal{W}$ to bound $\max_{w \in \mathcal{W}} \|w\|_2$ after the analogous Cauchy-Schwarz step as in \ref{afl}, we have
\begin{equation}
    \mathbb{E} \left[ \max_{w \in \mathcal{W}} \sum_{t=1}^T \langle w, \hat{g}_w^t - \nabla_w \phi(w^t, p^t) \rangle \right] \leq R_\mathcal{W} \sqrt{T}\tilde{\sigma}_w
\end{equation}
For the third and final term in \eqref{terms},
note that by the Law of Iterated Expectations and the unbiasedness of the stochastic gradients, we have that for any $t\geq 1$,
\begin{align}
  & \mathbb{E}[\langle p^t, (\nabla_p \phi(w^t,p^t) - \hat{g}_p^t)\rangle - \langle w^t, (\nabla_w \phi(w^t,p^t) - \hat{g}_w^t)\rangle ] \nonumber \\
    &= \mathbb{E}\left[\mathbb{E}\left[ \langle p^t,  (\nabla_p \phi(w^t,p^t) - \hat{g}_p^t) \rangle - \langle w^t, (\nabla_w \phi(w^t,p^t) - \hat{g}_w^t)\rangle| w^t,p^t \right]\right] \nonumber \\
    &= 0 \nonumber
\end{align}
Recalling \eqref{convexity} and \eqref{terms}, by combining the bounds on each of the terms and dividing by $T$, we obtain
\begin{equation}
    \mathbb{E} \left[ \max_{p \in \Delta_n} \phi(w^C_T, p) - \min_{w \in \mathcal{W} } \phi(w, p^C_T)  \right] \leq \dfrac{2 R_\mathcal{W}^2}{\eta_w T} + \dfrac{\eta_w \hat{G}_w^2}{2} +\dfrac{2}{\eta_p T} + \dfrac{\eta_p \hat{G}_p^2}{2} + \dfrac{R_\mathcal{W} \tilde{\sigma}_w}{\sqrt{T}} + \dfrac{\tilde{\sigma}_p}{\sqrt{T}}
\end{equation}
We minimize the above bound by setting the step sizes as
\begin{equation}
    \eta_w = \dfrac{2R_\mathcal{W}}{\hat{G}_w\sqrt{T}}, \quad \eta_p = \dfrac{2}{\hat{G}_p\sqrt{T}}
\end{equation}
to complete the proof, noting that $\tilde{\sigma}_w \leq \hat{G}_w$ and $\tilde{\sigma}_p \leq \hat{G}_p$.
\end{proof}

\section{Proof of Lemma \ref{lemma-strong}}
The result is a sample-approximation version of Theorem 1 in \citep{finn2019online}. We include our version of the proof here for completeness.
\begin{proof}
Note that $\hat{F}_i(w)$ is the empirical average of the functions $\hat{f}_{i,j}(w - \alpha \nabla \hat{f}_{i,j}(w, D^{\text{train}}_{i,j}), D^{\text{test}}_{i,j})$ for $j = 1,\dots,m_i$, so we can write $\hat{F}_i(w)$ as the empirical expectation over $j$:
\begin{equation}
\hat{\mathbb{E}}_j[\hat{f}_{i,j}(w - \alpha \nabla \hat{f}_{i,j}(w, D^{\text{train}}_{i,j}), D^{\text{test}}_{i,j})] \coloneqq \frac{1}{m_i} \sum_{j=1}^{m_i} \hat{f}_{i,j}(w - \alpha \nabla \hat{f}_{i,j}(w, D^{\text{train}}_{i,j}), D^{\text{test}}_{i,j}) = \hat{F}_i(w)
\end{equation} 
Using this notation, we show the strong convexity of $\hat{F}_i$ when $\alpha < 1/M$ and each $
\hat{f}_{i,j}(\cdot, D^{\text{test}}_{i,j})$ is $\mu$-strongly convex in addition to satisfying Assumption \ref{bound_grad}. We have
\begin{align}
   & \| \nabla \hat{F}_i(u) - \nabla \hat{F}_i(v) \| \nonumber \\
    &= \| \hat{\mathbb{E}}_{j} \big[(I - \alpha \nabla^2 \hat{f}_{i,j}(u, D^{\text{train}}_{i,j})) \nabla \hat{f}_{i,j}(u - \alpha \nabla \hat{f}_{i,j}(u, D^{\text{train}}_{i,j}), D^{\text{test}}_{i,j})\nonumber \\
    & \qquad \qquad - (I - \alpha \nabla^2 \hat{f}_{i,j}(v, D^{\text{train}}_{i,j})) \nabla \hat{f}_{i,j}(v - \alpha \nabla \hat{f}_{i,j}(v, D^{\text{train}}_{i,j}), D^{\text{test}}_{i,j}) \big] \| \nonumber \\
    &= \| \hat{\mathbb{E}}_{j} \big[(I - \alpha \nabla^2 \hat{f}_{i,j}(u, D^{\text{train}}_{i,j})) \big(\nabla \hat{f}_{i,j}(u - \alpha \nabla \hat{f}_{i,j}(u, D^{\text{train}}_{i,j}), D^{\text{test}}_{i,j}) \nonumber \\
    & \quad \quad \quad - \nabla \hat{f}_{i,j}(v - \alpha \nabla \hat{f}_{i,j}(v, D^{\text{train}}_{i,j}), D^{\text{test}}_{i,j})\big) - \bigg((I - \alpha \nabla^2 \hat{f}_{i,j}(v, D^{\text{train}}_{i,j})) \nonumber \\
    &\quad \quad \quad - (I - \alpha \nabla^2 \hat{f}_{i,j}(u, D^{\text{train}}_{i,j}))\big) \nabla \hat{f}_{i,j}(v - \alpha \nabla \hat{f}_{i,j}(v, D^{\text{train}}_{i,j}), D^{\text{test}}_{i,j}) \big] \| \label{smooth1} \\
    & \geq \| \hat{\mathbb{E}}_{j} \big[(I - \alpha \nabla^2 \hat{f}_{i,j}(u, D^{\text{train}}_{i,j})) (\nabla \hat{f}_{i,j}(u - \alpha \nabla \hat{f}_{i,j}(u, D^{\text{train}}_{i,j}), D^{\text{test}}_{i,j}) \nonumber \\
    & \quad \quad \quad - \nabla \hat{f}_{i,j}(v - \alpha \nabla \hat{f}_{i,j}(v, D^{\text{train}}_{i,j}), D^{\text{test}}_{i,j})) \big] \| \nonumber \\
    &\quad - \| \hat{\mathbb{E}}_{j}  \big[ ((I - \alpha \nabla^2 \hat{f}_{i,j}(v, D^{\text{train}}_{i,j})) \nonumber \\
    &\quad \quad \quad - (I - \alpha \nabla^2 \hat{f}_{i,j}(u, D^{\text{train}}_{i,j}))) \nabla \hat{f}_{i,j}(v - \alpha \nabla \hat{f}_{i,j}(v, D^{\text{train}}_{i,j}), D^{\text{test}}_{i,j}) \big] \| \nonumber \\
    &= \| \hat{\mathbb{E}}_{j} \big[(I - \alpha \nabla^2 \hat{f}_{i,j}(u, D^{\text{train}}_{i,j})) (\nabla \hat{f}_{i,j}(u - \alpha \nabla \hat{f}_{i,j}(u, D^{\text{train}}_{i,j}), D^{\text{test}}_{i,j}) \nonumber \\
    &\quad - \nabla \hat{f}_{i,j}(v - \alpha \nabla \hat{f}_{i,j}(v, D^{\text{train}}_{i,j}), D^{\text{test}}_{i,j})) \big] \| \nonumber \\
    &\quad - \alpha \| \hat{\mathbb{E}}_{j}  \big[ (\nabla^2 \hat{f}_{i,j}(u, D^{\text{train}}_{i,j}) - \nabla^2 \hat{f}_{i,j}(v, D^{\text{train}}_{i,j})) \nabla \hat{f}_{i,j}(v - \alpha \nabla \hat{f}_{i,j}(v, D^{\text{train}}_{i,j}), D^{\text{test}}_{i,j}) \big] \| \label{strcon} 
\end{align}
To lower bound the first term, we use the $\hat{M}$-smoothness of $\hat{f}_{i,j}(\cdot, D^{\text{train}}_{i,j})$, which implies that the minimum eigenvalue of $I - \alpha \nabla^2 \hat{f}_{i,j}(u, D^{\text{train}}_{i,j})$ is at least $1-\alpha \hat{M}$ for all $u \in \mathcal{W}$. Thus, 
\begin{align}
    \| \hat{\mathbb{E}}_{j} &\big[(I - \alpha \nabla^2 \hat{f}_{i,j}(u, D^{\text{train}}_{i,j})) (\nabla \hat{f}_{i,j}(u - \alpha \nabla \hat{f}_{i,j}(u, D^{\text{train}}_{i,j}), D^{\text{test}}_{i,j}) \nonumber \\
    &\quad \quad - \nabla \hat{f}_{i,j}(v - \alpha \nabla \hat{f}_{i,j}(v, D^{\text{train}}_{i,j}), D^{\text{test}}_{i,j})) \big] \| \nonumber \\
    &\geq (1-\alpha \hat{M})\|  \hat{\mathbb{E}}_{j} \big[\nabla \hat{f}_{i,j}(u - \alpha \nabla \hat{f}_{i,j}(u, D^{\text{train}}_{i,j}), D^{\text{test}}_{i,j}) - \nabla \hat{f}_{i,j}(v -\alpha \nabla \hat{f}_{i,j}(v, D^{\text{train}}_{i,j}), D^{\text{test}}_{i,j}) \big]  \| \label{str-con1}
\end{align}
By the $\hat{\mu}$-strong convexity of $\hat{f}_{i,j}(\cdot, D^{\text{test}}_{i,j})$ and the triangle inequality, we have
\begin{align}
\| \hat{\mathbb{E}}_{j} \big[\nabla \hat{f}_{i,j}(u - &\alpha \nabla \hat{f}_{i,j}(u, D^{\text{train}}_{i,j}), D^{\text{test}}_{i,j}) - \nabla \hat{f}_{i,j}(v -\alpha \nabla \hat{f}_{i,j}(v, D^{\text{train}}_{i,j}), D^{\text{test}}_{i,j}) \big]  \| \nonumber \\
&\geq \hat{\mu} \| \hat{\mathbb{E}}_{j} \big[u - \alpha \nabla \hat{f}_{i,j}(u, D^{\text{train}}_{i,j}) - (v -\alpha \nabla \hat{f}_{i,j}(v, D^{\text{train}}_{i,j})) \big]  \| \nonumber\\
&\geq \hat{\mu} \bigg(\|  u - v \| - \alpha\| \hat{ \mathbb{E}}_{j} \big[ \nabla \hat{f}_{i,j}(v, D^{\text{train}}_{i,j}) - \nabla \hat{f}_{i,j}(u, D^{\text{train}}_{i,j})\big]  \|\bigg) \nonumber\\
&\geq \hat{\mu} \left(\|  u - v \| - \alpha \hat{\mathbb{E}}_j\| \nabla \hat{f}_{i,j}(v, D^{\text{train}}_{i,j}) - \nabla \hat{f}_{i,j}(u, D^{\text{train}}_{i,j}) \|\right)\nonumber \\
&\geq \mu \left(\|  u - v \| - \alpha \hat{M}\|u-v\| \right)\label{str-con2}
\end{align}
where the second-to-last inequality follows from Jensen's Inequality and the last inequality follows from the $\hat{M}$-smoothness of each $\hat{f}_{i,j}(\cdot, D^{\text{train}}_{i,j})$
Next we upper bound the second term in \eqref{strcon}. We have
\begin{align}
\| \hat{\mathbb{E}}_{j}  &\big[ (\nabla^2 \hat{f}_{i,j}(u, D^{\text{train}}_{i,j}) - \nabla^2 \hat{f}_{i,j}(v, D^{\text{train}}_{i,j}) )\nabla \hat{f}_{i,j}(v - \alpha \nabla \hat{f}_{i,j}(v, D^{\text{train}}_{i,j}), D^{\text{test}}_{i,j}) \big] \| \nonumber\\
    &\leq  \hat{\mathbb{E}}_{j}  \big[ \|(\nabla^2 \hat{f}_{i,j}(u, D^{\text{train}}_{i,j}) - \nabla^2 \hat{f}_{i,j}(v, D^{\text{train}}_{i,j}) )\nabla \hat{f}_{i,j}(v - \alpha \nabla \hat{f}_{i,j}(v, D^{\text{train}}_{i,j}), D^{\text{test}}_{i,j}) \| \big] \label{omega-j} \\
    &\leq  \hat{\mathbb{E}}_{j}  \big[ \|(\nabla^2 \hat{f}_{i,j}(u, D^{\text{train}}_{i,j}) - \nabla^2 \hat{f}_{i,j}(v, D^{\text{train}}_{i,j}) )\| \|\nabla \hat{f}_{i,j}(v - \alpha \nabla \hat{f}_{i,j}(v, D^{\text{train}}_{i,j}), D^{\text{test}}_{i,j}) \| \big] \label{omega-cs} \\
    &\leq \sqrt{\hat{\mathbb{E}}_{j}  [ \|\nabla^2 \hat{f}_{i,j}(u, D^{\text{train}}_{i,j}) - \nabla^2 \hat{f}_{i,j}(v, D^{\text{train}}_{i,j})\|^2] \hat{\mathbb{E}}_{j} [ \|\nabla \hat{f}_{i,j}(v - \alpha \nabla \hat{f}_{i,j}(v, D^{\text{train}}_{i,j}), D^{\text{test}}_{i,j}) \|^2 ] } \label{omega-cs2} \\
    &\leq  \hat{L} \sqrt{\hat{\mathbb{E}}_{j}  [ \|\nabla^2 \hat{f}_{i,j}(u, D^{\text{train}}_{i,j}) - \nabla^2 \hat{f}_{i,j}(v, D^{\text{train}}_{i,j})\|^2]} \label{omega-l} \\
    &\leq \hat{L} \hat{\rho} \|u-v\| \label{omega-x}
\end{align}
where \eqref{omega-j} follows from Jensen's Inequality, \eqref{omega-cs} and \eqref{omega-cs2} follow from the Cauchy-Schwarz Inequality, \eqref{omega-l} follows from the $\hat{L}$-Lipschitzness of $\hat{f}_{i,j}(v, D^{\text{test}}_{i,j})$ for all $j \in [m_i]$, and \eqref{omega-x} follows from Assumption \ref{bound_grad}. 
Combining \eqref{strcon}, \eqref{str-con1}, and \eqref{str-con2} and \eqref{omega-x} yields that $\hat{F}_i$ is $\tilde{\mu} \coloneqq (\hat{\mu}(1-\alpha \hat{M})^2 - \alpha \hat{L} \hat{\rho}$)-strongly convex under the given conditions. 
\end{proof}

\section{Proof of Lemma \ref{lemma-smooth}}

\begin{proof}
We show the smoothness of each $\hat{F}_i$ by upper bounding the norm of the difference of its gradients. Using \eqref{smooth1} and the triangle inequality,
\begin{align}
   & \| \nabla \hat{F}_i(u) - \nabla \hat{F}_i(v) \| \nonumber \\
    &\leq \| \hat{\mathbb{E}}_{j} \big[(I - \alpha \nabla^2 \hat{f}_{i,j}(u, D^{\text{train}}_{i,j})) (\nabla \hat{f}_{i,j}(u - \alpha \nabla \hat{f}_{i,j}(u, D^{\text{train}}_{i,j}), D^{\text{test}}_{i,j}) - \nabla \hat{f}_{i,j}(v - \alpha \nabla \hat{f}_{i,j}(v, D^{\text{train}}_{i,j}), D^{\text{test}}_{i,j}))\big] \| \nonumber \\
    & \quad + \| \hat{\mathbb{E}}_j \big[((I - \alpha \nabla^2 \hat{f}_{i,j}(v, D^{\text{train}}_{i,j})) - (I - \alpha \nabla^2 \hat{f}_{i,j}(u, D^{\text{train}}_{i,j}))) \nabla \hat{f}_{i,j}(v - \alpha \nabla \hat{f}_{i,j}(v, D^{\text{train}}_{i,j}), D^{\text{test}}_{i,j}) \big] \| \label{two_terms}
\end{align}
We consider the two terms in the right hand side of \eqref{two_terms} separately. Denoting the first term as $\Xi$, we use Jensen's Inequality then the Cauchy-Schwarz Inequality twice, as in \eqref{omega-cs} and \eqref{omega-cs2}, to obtain
\begin{align}
    \Xi &\leq \big( \hat{\mathbb{E}}_{j} \big[ \|I - \alpha \nabla^2 \hat{f}_{i,j}(u, D^{\text{train}}_{i,j})\|^2 \big]  \hat{\mathbb{E}}_{j} \big[ \| \nabla \hat{f}_{i,j}(u - \alpha \nabla \hat{f}_{i,j}(u, D^{\text{train}}_{i,j}), D^{\text{test}}_{i,j}) \nonumber \\
    &\quad \quad \quad - \nabla \hat{f}_{i,j}(v - \alpha \nabla \hat{f}_{i,j}(v, D^{\text{train}}_{i,j}), D^{\text{test}}_{i,j}) \|^2 \big] \big)^{1/2} \nonumber \\
    &\leq (1+\alpha \hat{M})\sqrt{ \hat{\mathbb{E}}_{j} \big[  \| \nabla \hat{f}_{i,j}(u - \alpha \nabla \hat{f}_{i,j}(u, D^{\text{train}}_{i,j}), D^{\text{test}}_{i,j}) - \nabla \hat{f}_{i,j}(v - \alpha \nabla \hat{f}_{i,j}(v, D^{\text{train}}_{i,j}), D^{\text{test}}_{i,j}) \|^2 \big]  } \label{smooth-var}
\end{align}
where to obtain \eqref{smooth-var} we have used the $M$-smoothness of $\hat{f}_{i,j}$. Considering the term remaining inside the square root, we have 
\begin{align}
  \hat{\mathbb{E}}_{j} &\big[  \| \nabla \hat{f}_{i,j}(u - \alpha \nabla \hat{f}_{i,j}(u, D^{\text{train}}_{i,j}), D^{\text{test}}_{i,j}) - \nabla \hat{f}_{i,j}(v - \alpha \nabla \hat{f}_{i,j}(v, D^{\text{train}}_{i,j}), D^{\text{test}}_{i,j}) \|^2 \big] \nonumber \\
    &\leq \hat{M}^2 \hat{\mathbb{E}}_{j} \big[ \|u - \alpha \nabla \hat{f}_{i,j}(u, D^{\text{train}}_{i,j}) - (v - \alpha \nabla \hat{f}_{i,j}(v, D^{\text{train}}_{i,j}))\|^2 \big] \label{smooth-smooth} \\
    &= \hat{M}^2 \hat{\mathbb{E}}_{j} \big[ \|u - v\|^2 + 2\alpha(u-v)^T(\nabla \hat{f}_{i,j}(v, D^{\text{train}}_{i,j}) - \nabla \hat{f}_{i,j}(u, D^{\text{train}}_{i,j})) \nonumber\\
&\quad    + \alpha^2\| \nabla \hat{f}_{i,j}(u, D^{\text{train}}_{i,j}) - \nabla \hat{f}_{i,j}(v, D^{\text{train}}_{i,j})\|^2 \big] \nonumber \\
    &= \hat{M}^2 \bigg(\|u - v\|^2 + 2\alpha^2(u-v)^T \mathbb{E}_{j} [ \nabla \hat{f}_{i,j}(u, D^{\text{train}}_{i,j}) - \nabla \hat{f}_{i,j}(v, D^{\text{train}}_{i,j})] \nonumber\\
    &\qquad\qquad \qquad + \alpha^2 \mathbb{E}_{j} [\| \nabla \hat{f}_{i,j}(u, D^{\text{train}}_{i,j}) - \nabla \hat{f}_{i,j}(v, D^{\text{train}}_{i,j})\|^2 ]\bigg) \nonumber \\
    &\leq \hat{M}^2 \left(\|u-v\|^2 + 2 \alpha\hat{M} \|u-v\|^2 + \alpha^2 \hat{M}^2\|u-v\|^2\right) \label{smooth-smooth} \\
    &= \hat{M}^2 \left(1+\alpha \hat{M}\right)^2 \|u-v\|^2 \label{smooth-a}
\end{align}
where \eqref{smooth-smooth} follows from the $\hat{M}$-smoothness of $\hat{f}_{i,j}(\cdot, D^{\text{train}}_{i,j})$ and the Cauchy Schwarz Inequality. Thus we have
\begin{align}
    \Xi \leq \hat{M}(1+\alpha \hat{M})^2\|u-v\|
\end{align}
Note that we have already upper bounded the second term in \eqref{two_terms} in the previous lemma (see Equation \eqref{omega-x}). Thus we have that the smoothness parameter of $\hat{F}_i$ is 
\begin{equation}
    \tilde{M}_i \coloneqq \hat{M}(1+\alpha \hat{M})^2  + \alpha \hat{L} \hat{\rho}
\end{equation}
\end{proof}

\section{Proof of Lemma \ref{lemma_var}}\label{app:grad_var}
\begin{proof}
We again use the shorthand $X_k$ as defined in Appendix \eqref{defX}, and we also define
\begin{equation}
X_{i,j} =p_{i} (I - \alpha \nabla^2 \hat{f}_{i,j}(w, D_{i,j}^{\text{train}})) \nabla \hat{f}_{i,j}(w - \alpha \nabla \hat{f}_{i, j}(w, D_{i,j}^{\text{train}}), D_{i,j}^{\text{test}}) 
\end{equation}
for a fixed $i \in [n]$ and $j \in [m_i]$. Note that $X_k$ is a random variable while $X_{i,j}$ is deterministic.
Also observe that $\hat{g}_w(w,p) = \frac{1}{C}\sum_{k=1}^C nX_k$, that each $n X_k$ is an unbiased estimate of $g_w(w,p)$, and the $X_k$'s are independent.
Using these facts, we have
\begin{align}
\mathbb{E}[ \|  \hat{g}_w(w,p) - g_w(w,p) \|_2^2 ] &= \mathbb{E}[ \|  \frac{1}{C} \sum_{k =1}^C nX_k - g_w(w,p) \|_2^2 ] \\
&=  \frac{1}{C^2} \mathbb{E} [ \sum_{k =1}^C  \|  nX_k - g_w(w,p) \|_2^2 ]  \\
&= \frac{1}{C} \mathbb{E} [ \|  n X_1 - g_w(w,p) \|_2^2 ] \\
&=  \frac{1}{nC} \sum_{i=1}^n  \frac{1}{m_i} \sum_{j=1}^{m_i}  \|  nX_{i,j} - g_w(w,p) \|_2^2 \\
&= \frac{1}{nC} \sum_{i=1}^n  \frac{1}{m_i} \sum_{j=1}^{m_i}  \|  nX_{i,j} - np_i \nabla \hat{F}_i(w)\|_2^2 +  \|np_i \nabla \hat{F}_i(w)  - g_w(w,p) \|_2^2  \nonumber \\
&\quad - 2(nX_{i,j}- np_i \nabla \hat{F}_i(w))(np_i \nabla \hat{F}_i(w) - g_w(w,p))
\end{align}
Consider
\begin{align*}
\sum_{j=1}^{m_i} &(nX_{i,j}- np_i \nabla \hat{F}_i(w))(np_i \nabla \hat{F}_i(w) - g_w(w,p)) \\
&= (np_i \nabla \hat{F}_i(w) - g_w(w,p))\sum_{j=1}^{m_i} (nX_{i,j}- np_i \nabla \hat{F}_i(w)) \\
&=n (np_i \nabla \hat{F}_i(w) - g_w(w,p))\Bigg[\sum_{j=1}^{m_i} \left( p_{i} (I - \alpha \nabla^2 \hat{f}_{i}(w, D_{i,j}^{\text{train}})) \nabla \hat{f}_{i}(w - \alpha \nabla \hat{f}_{i}(w, D_{i,j}^{\text{train}}), D_{i,j}^{\text{test}})\right) 
\nonumber\\
&\qquad\qquad\qquad\qquad\qquad\qquad\qquad\qquad- m_i p_i \nabla \hat{F}_i(w)\Bigg] \\
&= n (np_i \nabla \hat{F}_i(w) - g_w(w,p))\left[m_i p_i \nabla \hat{F}_i(w) - m_i p_i \nabla \hat{F}_i(w) \right] \\
&= 0
\end{align*}
Therefore we have
\begin{align*}
&\mathbb{E}[ \|  \hat{g}_w(w,p) - g_w(w,p) \|_2^2 ]\\
 &=\frac{1}{nC} \sum_{i=1}^n  \frac{1}{m_i} \sum_{j=1}^{m_i}  \|  nX_{i,j} - np_i \nabla \hat{F}_i(w)\|_2^2 +  \|np_i \nabla \hat{F}_i(w)  - g_w(w,p) \|_2^2  \nonumber \\
&= \frac{n}{C} \sum_{i=1}^n  \frac{1}{m_i} \sum_{j=1}^{m_i}  \|  p_i X_{i,j} - p_i \sum_{j'=1}^{m_i} X_{i,j'}\|_2^2 +  \frac{n}{C} \sum_{i=1}^n  \|p_i \nabla \hat{F}_{i}(w)  - \frac{1}{n} \sum_{i'=1}^n p_{i'} \nabla \hat{F}_{i'}(w) \|_2^2  \nonumber \\
&=  \frac{n}{C} \sum_{i=1}^n \sigma_i^2+  \frac{n}{C}\sigma^2
\end{align*}
\end{proof}



\section{Proof of Theorem \ref{noncon-cor1}}

\begin{proposition} \label{noncon-lemma-1}
Suppose Assumption \ref{bound_grad} holds and $\mathcal{W} = \mathbb{R}^d$.
Let $\eta_w^t$ and $\eta_p^t$ be constant over all $t$, denoted by $\eta_w$ and $\eta_p$, respectively, where $\eta_w < ({2}/{\tilde{M}})$. Let $(w_T^\tau, p_T^\tau)$ be the solution returned by Algorithm \ref{alg1} after $T$ iterations.
Then, 
\begin{align}
     &\mathbb{E}[\|\nabla_w \phi(w_T^\tau, p_T^\tau)\|_2^2] 
     \leq \dfrac{2(\phi(w^1, p^1) + B)}{T(2\eta_w - {\eta_w^2\tilde{M}})} + \dfrac{4\eta_p \sqrt{n} B\hat{G}_p}{ (2\eta_w - {\eta_w^2\tilde{M}}) } 
     + \dfrac{\eta_w \tilde{M} \sigma_w^2}{(2 - \eta_w\tilde{M})}, \nonumber \\
  &  \mathbb{E}\left[ \phi(w_T^\tau, p_T^\tau)  \right] \geq \max_{p \in \Delta_n} \left\{ \mathbb{E} \left[ \phi(w^\tau_T, p)  \right] \right\} - \dfrac{1}{\eta_p T}- \dfrac{\eta_p  \hat{G}_p^2}{2} \nonumber
\end{align}
where $\hat{G}_p^2 = n(n+C-1) \hat{B}^2/ C$.
\end{proposition} 

\begin{proof}
Note that
\begin{align}
    \mathbb{E}[\|\nabla_w \phi(w_T^\tau, p_T^\tau)\|_2^2] &= \mathbb{E}\left[\mathbb{E}_{\tau}[\|\nabla_w \phi(w_T^\tau, p_T^\tau)\|_2^2]\right] \label{abc} \\
    &= \mathbb{E}\left[\dfrac{1}{T}\sum_{t=1}^T\|\nabla_w \phi(w^t, p^t)\|_2^2\right] \label{abcd} \\
    &= \dfrac{1}{T} \sum_{t=1}^T \mathbb{E} \left[\|g_w^{t}\|^2 \right] \label{abcde}
\end{align}
where the un-subscripted expectation in the right hand sides of \eqref{abc} and \eqref{abcd} is over the stochastic gradients which determine the sequence $\{(w^t,p^t)\}_t$.
Thus to show the bound on $ \mathbb{E}[\|\nabla_w \phi(w_T^\tau, p_T^\tau)\|_2^2]$ in Proposition \ref{noncon-lemma-1}, we bound the right hand side of \eqref{abcde}.
To do so we borrow ideas from the proof of Theorem 1 in \citet{qian2019robust}. First recall that by Lemma \ref{lemma-smooth}, $\hat{F}_i$ is $\tilde{M}$-smooth for each $i \in \{1,...,n\}$. Then for any $u,v \in \mathcal{W}$,
\begin{equation}
    \hat{F}_i(u) \leq \hat{F}_i(v) + \nabla \hat{F}_i(v)^T (u-v) + \dfrac{\tilde{M}}{2} \|u-v \|^2 \label{smoothF}
\end{equation}
Conditioned on the history up to iteration $t$, denoted by $\mathcal{F}^t$, the above equation implies
\begin{align}
  &  \mathbb{E} \left[ \sum_{i=1}^n p^t_{i} \hat{F}_i(w^{t+1}) | \mathcal{F}^t \right] 
\nonumber\\  &\leq \mathbb{E} \left[ \sum_{i=1}^n p^t_{i} \hat{F}_i(w^{t}) + \left(\nabla_w \sum_{i=1}^n p^t_{i}\hat{F}_i(w^t)\right)^T (w^{t+1}-w^t) + \dfrac{\tilde{M}}{2} \|w^{t+1} - w^t \|^2 | \mathcal{F}^t \right]
\end{align}
Note that $\nabla_w \sum_{i=1}^n p^t_{i}\hat{F}_i(w^t) = g^{t}_w$ and $w^{t+1} - w^t = -\eta_w \hat{g}^{t}_w$. Thus, we have
\begin{align}
    \mathbb{E} \left[ \sum_{i=1}^n p^t_{i} \hat{F}_i(w^{t+1}) | \mathcal{F}^t \right] &\leq \mathbb{E} \left[ \sum_{i=1}^n p^t_{i} \hat{F}_i(w^{t}) - \eta_w (g^{t}_w)^T \hat{g}^{t}_w + \dfrac{\tilde{M}}{2} \eta_w^2 \|\hat{g}_w^{t} \|^2 | \mathcal{F}^t \right] \\
    &= \sum_{i=1}^n p^t_{i} \hat{F}_i(w^{t})  - \eta_w  \|g_w^{t}\|^2 + \dfrac{\tilde{M}}{2} \eta_w^2\left(\|g_w^{t} \|^2 + \mathbb{E} \left[  \|\hat{g}_w^t - g_w^{t} \|^2 | \mathcal{F}^t \right]\right)\label{second}
\end{align}
where \eqref{second} follows because $\hat{g}_w^{t}$ is an unbiased estimate of $g_w^{t}$. Using Lemma \ref{lemma_var}, we have
\begin{equation}
    \mathbb{E} \left[ \sum_{i=1}^n p^t_{i} \hat{F}_i(w^{t+1}) | \mathcal{F}^t \right] \leq  \sum_{i=1}^n p^t_i \hat{F}_i(w^{t}) -\left(\eta_w - \dfrac{\eta_w^2\tilde{M}}{2}\right)\|g_w^{t}\|^2 + \dfrac{\eta_w^2 \tilde{M}  \sigma^2_w}{2} 
\end{equation}
Rearranging the terms, we obtain
\begin{align}
    \left(\eta_w - \tfrac{\eta_w^2\tilde{M}}{2}\right) \|g_w^{t}\|^2  &\leq \mathbb{E} \left[ \sum_{i=1}^n p_{i}^t \hat{F}_i(w^{t}) - \sum_{i=1}^n p_{i}^t \hat{F}_i(w^{t+1}) | \mathcal{F}^t  \right] + \dfrac{\eta^2_w \tilde{M}  \sigma_w^2}{2} \\
    &= \mathbb{E} \left[ \sum_{i=1}^n p^t_i \hat{F}_i(w^{t}) - \sum_{i=1}^n p^{t+1}_i \hat{F}_i(w^{t+1}) | \mathcal{F}^t \right] \\
    &\quad + \mathbb{E} \left[ \sum_{i=1}^n p^{t+1}_{i} \hat{F}_i(w^{t+1}) - \sum_{i=1}^n p^{t}_i \hat{F}_i(w^{t+1}) | \mathcal{F}^t \right] + \dfrac{\eta_w^2 \tilde{M}  \sigma_w^2}{2} \label{noncon-tosum}
\end{align}
We bound the second expectation in the above equation:
\begin{align}
    \mathbb{E} \left[ \sum_{i=1}^n p^{t+1}_{i} \hat{F}_i(w^{t+1}) - \sum_{i=1}^n p^{t}_{i} \hat{F}_i(w^{t+1}) | \mathcal{F}^t \right] &= \mathbb{E} \left[ \sum_{i=1}^n (p^{t+1}_{i} - p^{t}_{i}) \hat{F}_i(w^{t+1}) | \mathcal{F}^t \right] \nonumber \\
    &\leq \mathbb{E} \left[ \|p^{t+1} - p^{t} \|_2 \sum_{i=1}^n \left(\hat{F}_i(w^{t+1})\right)^{1/2} | \mathcal{F}^t \right] \label{noncon-cs} \\
    &\leq \sqrt{n}\hat{B} \mathbb{E} \left[ \|p^{t+1} - p^{t} \|_2 | \mathcal{F}^t \right] \label{noncon-p} \\
    &\leq 2\sqrt{n}\hat{B} (\mathbb{E} \left[ \|\eta_p \hat{g}_p^t \|_2 | \mathcal{F}^t  \right])\label{noncon-t} \\
    &= 2\eta_p\sqrt{n}\hat{B}\hat{G}_p\label{noncon-g}
\end{align}
where \eqref{noncon-cs} follows from the Cauchy-Shwarz Inequality, \eqref{noncon-p} follows by the bound on $\hat{f}_{i,j}$ for all $i$, \eqref{noncon-t} follows from the update rule for $p$ combined with the projection property (since $p^t \in \Delta_n$, $\|p^t - (p^t + \eta_p \hat{g}^t_p) \| \geq \|p^t - \Pi_{\Delta_n}(p^t + \eta_p \hat{g}^t_p)\|$), and \eqref{noncon-g} follows by Lemma \ref{bound_grad}, noting $\hat{G}_p^2 = \frac{n(n+C-1) \hat{B}}{C}$.
Using this result, summing \eqref{noncon-tosum} from $t=1$ to $T$, and taking the expectation over all the stochastic gradients of both sides and using the Law of Iterated Expectations to remove the conditioning on $\mathcal{F}^t$, we obtain
\begin{align}
   & \left(\eta_w - \tfrac{\eta_w^2\tilde{M}}{2}\right) \sum_{t=1}^T \mathbb{E}\left[\|g_w^{t}\|^2 \right] 
  \nonumber\\ &\leq \mathbb{E} \left[ \sum_{i=1}^n p^1_{i} \hat{F}_i(w^{1}) \right] - \mathbb{E}\left[ \sum_{i=1}^n p^{T+1}_{i} \hat{F}_i(w^{T+1}) \right] + 2T\eta_p\sqrt{n}\hat{B}\hat{G}_p + \dfrac{T \tilde{M} \eta^2_w \sigma_w^2}{2} \nonumber \\
    &\leq \phi(w^1,p^1) + \hat{B} + 2T\eta_p \sqrt{n} \hat{B}\hat{G}_p + \dfrac{T\eta^2_w \tilde{M}  \sigma_w^2}{2} \nonumber
\end{align}
Next, dividing both sides by $T\left(\eta_w - \tfrac{\eta_w^2\tilde{M}}{2}\right)$  we have
\begin{align}
    \dfrac{1}{T} \sum_{t=1}^T \mathbb{E} \left[\|g_w^{t}\|^2 \right] &\leq \dfrac{\phi(w^1, p^1) + \hat{B}}{T\left(\eta_w - \tfrac{\eta_w^2\tilde{M}}{2}\right)} + \dfrac{2\eta_p \sqrt{n} \hat{B}\hat{G}_p}{ \left(\eta_w - \tfrac{\eta_w^2\tilde{M}}{2}\right) } + \dfrac{\eta_w \tilde{M} \sigma_w^2}{\left(2 - \eta_w\tilde{M}\right)} 
\end{align}
which by \eqref{abcde} is the desired bound on $\mathbb{E}[\|\nabla_w \phi (w_T^\tau, p_T^\tau)\|_2^2]$.

Next we show the bound on the optimality of $p^\tau_T$. As before, we start by evaluating the expectation over $\tau$:
\begin{align}
   \mathbb{E}\left[ \phi(w_T^\tau, p_T^\tau)  \right] &= \mathbb{E} \left[ \mathbb{E}_\tau \left[ \phi(w_T^\tau, p_T^\tau)  \right] \right] \\
   &= \mathbb{E} \left[ \dfrac{1}{T}\sum_{t=1}^T \phi(w^t, p^t) \right] \\
   &= \dfrac{1}{T}\sum_{t=1}^T \mathbb{E} \left[ \phi(w_T^\tau, p_T^\tau) \right] \label{noncon-p-equal}
\end{align}
Next, since $\phi(w,p)$ is linear in $p$, we have that for any $p \in \Delta_n$ and any $t \in \{1,...,T\}$, 
\begin{align}
    \mathbb{E} \left[ \phi(w^t, p) - \phi(w^t, p^t)  | \mathcal{F}^t \right] &= \mathbb{E} \left[ (p-p^t) g_p^t  | \mathcal{F}^t \right] \nonumber \\
    &= \mathbb{E} \left[ (p-p^t) \hat{g}_p^t  | \mathcal{F}^t \right] + \mathbb{E} \left[ (p-p^t)(g_p^t - \hat{g}_p^t)  | \mathcal{F}^t  \right] \\
    &= \mathbb{E} \left[ (p-p^t) \hat{g}_p^t | \mathcal{F}^t \right] \label{noncon-p-hat}
\end{align}
where \eqref{noncon-p-hat} follows because $\hat{g}_p^t$ is an unbiased estimate of $g_p^t$. Using \eqref{noncon-p-hat} and the identity $2ab = a^2 + b^2 - (a-b)^2$ with $a = p-p^t$ and $b = \eta_p \hat{g}_p^t$ yields
\begin{align}
    \mathbb{E} \left[ \phi(w^t, p) - \phi(w^t, p^t) | \mathcal{F}^t \right] &= \mathbb{E} \left[\dfrac{1}{2\eta_p} \left(\| p-p^t  \|_2^2 + (\eta_p)^2\|\hat{g}_p^t\|_2^2 - \|p - (p^t + \eta_p \hat{g}_p^t)  \|_2^2  \right) | \mathcal{F}^t  \right] \\
    &\leq \mathbb{E} \left[\dfrac{1}{2\eta_p} \left(\| p-p^t  \|_2^2 + (\eta_p)^2\|\hat{g}_p^t\|_2^2 - \|p - p^{t+1}  \|_2^2  \right) | \mathcal{F}^t  \right] \label{noncon-p-projprop} \\
    &\leq \mathbb{E} \left[\dfrac{1}{2\eta_p} \left(\| p-p^t  \|_2^2 + (\eta_p)^2\hat{G}_p^2 - \|p - p^{t+1}  \|_2^2  \right) | \mathcal{F}^t  \right] \label{noncon-p-gbound}
\end{align}
where \eqref{noncon-p-projprop} follows from the projection property and \eqref{noncon-p-gbound} follows from Lemma \ref{bound_grad}. Summing from $t = 1$ to $T$ and taking the expectation over all the stochastic gradients of both sides and using the Law of Iterated Expectations to remove the conditioning on $\mathcal{F}^t$,  we obtain
\begin{align}
    \sum_{t=1}^T \mathbb{E} \left[ \phi(w^t, p) - \phi(w^t, p^t) \right] &\leq \sum_{t=1}^T \dfrac{1}{2 \eta_p}\mathbb{E} \left[ \| p-p^t  \|_2^2\right] - \dfrac{1}{2\eta_p}\mathbb{E} \left[\|p - p^{t+1}  \|_2^2 \right] + \dfrac{\eta_p}{2}\hat{G}_p^2 \label{non-tele}\\
    &= \dfrac{1}{2\eta_p}\mathbb{E} \left[\|p - p^{1}  \|_2^2 \right] + \dfrac{\eta_p}{2} T \hat{G}_p^2 \\
    &\leq \dfrac{1}{\eta_p} + \dfrac{\eta_p T \hat{G}_p^2}{2} \label{noncon-p-pbound}
\end{align}
where \eqref{non-tele} follows from the telescoping sum and \eqref{noncon-p-pbound} follows from the fact that $p, p^1 \in \Delta_n$ and $\Delta_n$ is contained in an $\ell_2$ ball of radius 1. Dividing both sides of \eqref{noncon-p-pbound} by $T$ and rearranging terms
\begin{align}
    \dfrac{1}{T}\sum_{t=1}^T \mathbb{E} \left[ \phi(w^t, p^t) \right] &\geq \mathbb{E} \left[ \phi(w^\tau_T, p) \right] - \left(\dfrac{1}{\eta_p T}+ \dfrac{\eta_p  \hat{G}_p^2}{2}\right) \label{premax}
\end{align}
Finally, since \eqref{premax} holds for all $p \in \Delta_n$, we maximize the right hand side over $p \in \Delta_n$, yielding
\begin{align}
    \dfrac{1}{T}\sum_{t=1}^T \mathbb{E} \left[ \phi(w^t, p^t) \right] &\geq \max_{p \in \Delta_n} \left[ \phi(w^\tau_T, p)  \right] - \left(\dfrac{1}{\eta_p T}+ \dfrac{\eta_p  \hat{G}_p^2}{2}\right) \nonumber
\end{align}
From \eqref{noncon-p-equal}, the left hand side above is equal to $\mathbb{E}\left[ \phi(w_T^\tau, p_T^\tau)  \right]$, thus completing the proof.
\end{proof}

Theorem \ref{noncon-cor1} follows immediately from Proposition \ref{noncon-lemma-1} by setting the step sizes appropriately.

\section{Proof of Theorem \ref{noncon-thrm3}}

First we have the following proposition for unspecified constant stepsizes.
\begin{proposition} \label{noncon-compact}
Suppose Assumptions \ref{bound_grad} and \ref{assump_hess} hold and $\mathcal{W}$ is convex and compact.
Let the step sizes $\eta_w^t$ and $\eta_p^t$ be constant over all $t$, denoted by $\eta_w$ and $\eta_p$, respectively, where $\eta_w < ({2}/{\tilde{M}})$. Let $(w_T^\tau, p_T^\tau)$ be the solution returned by Algorithm \ref{alg1} after $T$ iterations.
Then we have 
\begin{align}
 &    \mathbb{E}[\|\bar{g}_w(w_T^\tau, p_T^\tau)\|_2^2] \leq \dfrac{2(\phi(w^1, p^1) + \hat{B})}{T(2\eta_w - {\eta_w^2\tilde{M}})} + \dfrac{4\eta_p \sqrt{n} \hat{B}\hat{G}_p}{ (2\eta_w - {\eta_w^2\tilde{M}}) } + \dfrac{\sigma_w^2}{(2 - \eta_w\tilde{M})}, \nonumber \\
    &\mathbb{E}\left[ \phi(w_T^\tau, p_T^\tau)  \right] \geq \max_{p \in \Delta_n} \left\{ \mathbb{E} \left[ \phi(w^\tau_T, p)  \right] \right\} - \dfrac{1}{\eta_p T}- \dfrac{\eta_p  \hat{G}_p^2}{2}. \nonumber
\end{align}
\end{proposition}

\begin{proof}
Here it is helpful to rewrite $\Pi_\mathcal{W}$ as a prox operation.
Defining $I_{\mathcal{W}}: \mathcal{W} \rightarrow \{0,+\infty\}$ as $I_{\mathcal{W}}(w) = 0$ if $w \in \mathcal{W}$ and $I_{\mathcal{W}}(w) = +\infty$ otherwise, the update rule for $w$ becomes:
\begin{align}
    w^{t+1} = \Pi_{\mathcal{W}}(w^t - \eta_w^t g_w^t)  = \argmin_{u \in \mathbb{R}^d} \{\langle \hat{g}_w^t, u\rangle + \dfrac{1}{2\eta_w^t} \|u\!-\!w^t\|_2^2 + I_\mathcal{W}(u)\}  = \text{prox}_{\eta_w^t I_{\mathcal{W}}}(w^t - \eta_w^t g_w^t)\nonumber \end{align}
and the projected stochastic gradient is equivalent to
\begin{equation}
    \bar{g}_w^t = \dfrac{1}{\eta_w^t}(w^t -  \text{prox}_{\eta_w^t I_{\mathcal{W}}}(w^t - \eta_w^t \hat{g}_w^t)  ) \nonumber
\end{equation}
The rewritten objective, using $I_{\mathcal{W}}$ to remove the constraint on $w$, is as follows:
\begin{equation}
\min_{w \in \mathbb{R}^d} \max_{p \in \Delta_n} \{ \Phi(w,p) \coloneqq \phi(w,p) + I_{\mathcal{W}}(w)\}
\end{equation}

With these notations in hand, we are ready to begin the proof.
We make analogous initial arguments to those in the proof of Theorem~2 in \citep{ghadimi2016mini}, and cite two results on the properties of the prox operation from the same paper. By the $\tilde{M}$-smoothness of $\hat{F}_i$ for each $i$, we have equation \eqref{smoothF}, and thus for any $t \in \{1,...,T\}$,
\begin{align}
    \sum_{i=1}^n p^t_{i} \hat{F}_i(w^{t+1}) &\leq \sum_{i=1}^n p^t_{i} \hat{F}_i(w^{t}) + \left(\nabla_w \sum_{i=1}^n p^t_{i}\hat{F}_i(w^t)\right)^T (w^{t+1}-w^t) + \dfrac{\tilde{M}}{2} \|w^{t+1} - w^t \|_2^2 \nonumber \\
    &= \sum_{i=1}^n p^t_{i} \hat{F}_i(w^{t}) - \eta^t_w\left(\nabla_w \sum_{i=1}^n p^t_{i}\hat{F}_i(w^t)\right)^T \bar{g}^t_w+ \dfrac{\tilde{M}}{2} (\eta_w^t)^2\|\bar{g}^t_w\|_2^2 \nonumber \\
    &= \sum_{i=1}^n p^t_{i} \hat{F}_i(w^{t}) - \eta^t_w\left(\hat{g}_w^t\right)^T \bar{g}^t_w+ \dfrac{\tilde{M}}{2} (\eta_w^t)^2\|\bar{g}^t_w\|_2^2 + \eta^t_w (\delta^t_w)^T \bar{g}^t_w\nonumber
\end{align}
where in the identity we have used the definitions of $\bar{g}^t$ and $\delta^t_w$. Next, using Lemma 1 in \citep{ghadimi2016mini} with $x = w^t$, $\gamma = \eta_w^t$, and $g = \hat{g}_w^t$, we obtain
\begin{align}
    \sum_{i=1}^n p^t_{i} \hat{F}_i(w^{t+1}) &\leq \sum_{i=1}^n p^t_{i} \hat{F}_i(w^{t}) - [\eta^t_w \|\bar{g}^t\|_2^2 + I_{\mathcal{W}}(w^{t+1}) - I_{\mathcal{W}}(w^t)] + \dfrac{\tilde{M}}{2} (\eta_w^t)^2\|\bar{g}^t_w\|_2^2 + \eta^t_w (\delta^t_w)^T \bar{g}^t_w\nonumber \\
    &= \sum_{i=1}^n p^t_{i} \hat{F}_i(w^{t}) - [\eta^t_w \|\bar{g}^t\|_2^2 + I_{\mathcal{W}}(w^{t+1}) - I_{\mathcal{W}}(w^t)] + \dfrac{\tilde{M}}{2} (\eta_w^t)^2\|\bar{g}^t_w\|_2^2 \nonumber \\
    & \quad \quad  \quad \quad + \eta^t_w (\delta^t_w)^T g^t + \eta^t_w (\delta^t_w)^T (\bar{g}^t_w- g^t) \nonumber
\end{align}
where $\delta_w^t \coloneqq \hat{g}_w^t - g_w^t$ and $g^t \coloneqq \dfrac{1}{\eta_w^t}(w^t -  \text{prox}_{\eta_w^t I_{\mathcal{W}}}(w^t - \eta_w^t  g_w^t)  )$ is the projected full gradient with respect to $w$. Thus after rearranging terms,
\begin{align}
    \Phi(w^{t+1}, p^{t}) &\leq \Phi(w^t, p^t) - \left(\eta_{w}^t - \dfrac{\tilde{M}}{2}(\eta_{w}^t)^2\right) \| \bar{g}^t_w\|^2_2 + \eta_w^t \langle \delta^t_w, g^t \rangle + \eta_w^t \|\delta^t_w\| \| \bar{g}^t_w- g^t \| \nonumber\\
    &\leq \Phi(w^t, p^t) - \left(\eta_{w}^t - \dfrac{\tilde{M}}{2}(\eta_{w}^t)^2\right) \| \bar{g}^t_w\|^2_2 + \eta_w^t \langle \delta^t_w, g^t \rangle + \eta_w^t \|\delta^t_w\|^2 \nonumber
\end{align}
where the last inequality follows from Proposition 1 in \citep{ghadimi2016mini} with $x = w^t$, $\gamma = \eta_w^t$, $g_1 = \hat{g}_w^t$, and $g_2 = g_w^t$. Rearranging terms, we have
\begin{align}
  & \left(\eta_{w}^t - \dfrac{\tilde{M}}{2}(\eta_{w}^t)^2\right) \| \bar{g}^t_w\|^2_2 \nonumber\\
  &\leq \Phi(w^t, p^t) - \Phi(w^{t+1}, p^{t}) + \eta_w^t \langle \delta^t_w, g^t \rangle + \eta_w^t \|\delta^t_w\|^2 \nonumber \\
   &= \left(\Phi(w^t, p^t) - \Phi(w^{t+1}, p^{t+1})\right) + \left(\Phi(w^{t+1}, p^{t+1}) - \Phi(w^{t+1}, p^{t})\right) + \eta_w^t \langle \delta^t_w, g^t \rangle + \eta_w^t \|\delta^t_w\|^2 \nonumber \\
   &= \left(\Phi(w^t, p^t) - \Phi(w^{t+1}, p^{t+1})\right) + \left(\phi(w^{t+1}, p^{t+1}) - \phi(w^{t+1}, p^{t})\right) + \eta_w^t \langle \delta^t_w, g^t \rangle + \eta_w^t \|\delta^t_w\|^2 \nonumber
\end{align}
Taking the expectation with respect to the stochastic gradients conditioned on the history up to time $t$ of each side, we have
\begin{align}
 & \left(\eta_{w}^t - \dfrac{\tilde{M}}{2}(\eta_{w}^t)^2\right) \mathbb{E} \left[\| \bar{g}^t_w\|^2_2 |\mathcal{F}^t \right]\nonumber\\
  &\leq \mathbb{E}\left[\left(\Phi(w^t, p^t) - \Phi(w^{t+1}, p^{t+1})\right) |\mathcal{F}^t \right] + \mathbb{E}\left[\left(\phi(w^{t+1}, p^{t+1}) - \phi(w^{t+1}, p^{t})\right) |\mathcal{F}^t  \right] \nonumber \\
    &\quad + \eta_w^t \mathbb{E} \left[ \langle \delta^t_w, g^t \rangle |\mathcal{F}^t \right] + \eta_w^t \mathbb{E} \left[\|\delta^t_w\|^2 |\mathcal{F}^t \right] \nonumber \\
    &= \mathbb{E}\left[\left(\Phi(w^t, p^t) - \Phi(w^{t+1}, p^{t+1})\right) |\mathcal{F}^t  \right] + \mathbb{E} \left[ \sum_{i=1}^n (p^{t+1}_{i} - p^{t}_{i} )\hat{F}_i(w^{t+1}) |\mathcal{F}^t \right] \nonumber \\
    &\quad + \eta_w^t \mathbb{E} \left[ \langle \delta^t_w, g^t \rangle |\mathcal{F}^t \right] + \eta_w^t \mathbb{E} \left[\|\delta^t_w\|^2 |\mathcal{F}^t  \right] \label{above}
\end{align}
Note that we can use the Holder Inequality to bound the second expectation in \eqref{above}. In doing so we obtain
\begin{align}
  & \left(\eta_{w}^t - \dfrac{\tilde{M}}{2}(\eta_{w}^t)^2\right) \mathbb{E} \left[\| \bar{g}^t_w\|^2_2 |\mathcal{F}^t \right]\nonumber\\  
  &\leq \mathbb{E}\left[\left(\Phi(w^t, p^t) - \Phi(w^{t+1}, p^{t+1})\right) |\mathcal{F}^t  \right] + \mathbb{E} \left[\|p^{t+1}-p^{t}\|_2 \left(\sum_{i=1}^n \hat{F}_i(w^{t+1})^2\right)^{1/2} |\mathcal{F}^t  \right] \nonumber \\    
    &\quad + \eta_w^t \mathbb{E} \left[ \langle \delta^t_w, g^t \rangle |\mathcal{F}^t  \right] + \eta_w^t \mathbb{E} \left[\|\delta^t_w\|^2 |\mathcal{F}^t  \right] \nonumber \\
    &\leq \mathbb{E}\left[\left(\Phi(w^t, p^t) - \Phi(w^{t+1}, p^{t+1})\right) |\mathcal{F}^t \right] + 2\sqrt{n}B\mathbb{E} \left[\|\eta_{p}^t \hat{g}_p^t\|_2 |\mathcal{F}^t \right] + \eta_w^t \mathbb{E} \left[ \langle \delta^t_w, g^t \rangle |\mathcal{F}^t  \right] \nonumber \\
    & \quad \quad \quad \quad + \eta_w^t \mathbb{E} \left[\|\delta^t_w\|^2_2 |\mathcal{F}^t \right] \label{uncon-b} \\
     &\leq \mathbb{E}\left[\left(\Phi(w^t, p^t) - \Phi(w^{t+1}, p^{t+1})\right) |\mathcal{F}^t \right] + 2\sqrt{n}B\eta_{p}^t \hat{G}_p + \eta_w^t \mathbb{E} \left[ \langle \delta^t_w, g^t \rangle |\mathcal{F}^t  \right] + \eta_w^t \mathbb{E} \left[\|\delta^t_w\|^2_2 |\mathcal{F}^t  \right] \label{uncon-d} \\
      &\leq \mathbb{E}\left[\left(\Phi(w^t, p^t) - \Phi(w^{t+1}, p^{t+1})\right) |\mathcal{F}^t \right] + 2\sqrt{n}B\eta_{p}^t \hat{G}_p + \eta_w^t \mathbb{E} \left[\|\delta^t_w\|^2_2 |\mathcal{F}^t  \right]  \label{zero} \\
      &\leq \mathbb{E}\left[\left(\Phi(w^t, p^t) - \Phi(w^{t+1}, p^{t+1})\right) |\mathcal{F}^t \right] + 2\sqrt{n}B\eta_{p}^t \hat{G}_p + \eta_w^t \sigma_w^2 \label{sigma}
\end{align}
where \eqref{uncon-b} follows from the definition of $B$ and the update rule for $p$ combined with the projection property, \eqref{uncon-d} follows from the definition of $\hat{G}_p$, \eqref{zero} follows from the facts that $g^{t}$ is a deterministic function of the stochastic samples that determine the stochastic gradients up to time $t$ and $\hat{g}_w^t$ is an unbiased estimate of $g^t_w$, and \eqref{sigma} follows from the computation of $\mathbb{E}[\|\delta_w\|^2]$ given in Lemma \ref{lemma_var}. Summing over $t=1,...,T$, setting the step sizes to be constants, and taking the expectation with respect to all of the stochastic gradients and using the Law of Iterated Expectations, we find
\begin{align}
    \left(\eta_{w} - \dfrac{\tilde{M}}{2}(\eta_{w})^2\right) \sum_{t=1}^T \mathbb{E} \left[\| \bar{g}^t_w\|^2_2  \right]
    &\leq \Phi(w^{1}, p^{1}) - \mathbb{E}\left[\Phi(w^{T+1}, p^{T+1}) \right] + 2T \eta_p B\sqrt{n} \hat{G}_p + T \eta_w \sigma_w^2 \nonumber \\
    &\leq \Phi(w^{1}, p^{1}) + B + 2T \eta_p B\sqrt{n} \hat{G}_p + T \eta_w \sigma_w^2 \nonumber
\end{align}
Next we divide both sides by $T\left(\eta_w -  \tfrac{\tilde{M}}{2}(\eta_{w})^2\right)$ to yield
\begin{align}
    \dfrac{1}{T} \sum_{t=1}^T \mathbb{E} \left[\| \bar{g}^t_w\|^2_2  \right]
    &\leq \dfrac{2(\phi(w^1, p^1) + B)}{T(2\eta_w - {\eta_w^2\tilde{M}})} + \dfrac{4\eta_p \sqrt{n} B\hat{G}_p}{ (2\eta_w - {\eta_w^2\tilde{M}}) } + {{ \dfrac{ \sigma_w^2}{(2 - \eta_w\tilde{M})}}} \nonumber
\end{align}
Using an analogous argument as \eqref{abcde}, we have that the left hand side of the above equation is equal to $\mathbb{E}[\|\bar{g}_w(w^\tau_T, p^\tau_T)\|^2_2]$, thus we have completed the proof of the convergence result in $w$.

For the convergence with respect to $p$, note that the update rule for $p^{t+1}$ is identical to the update rule analyzed in Proposition \ref{noncon-lemma-1}, and the output procedure is the same for both algorithms. Furthermore, since the convergence analysis of $p$ does not depend on the update rule for $w$,
the analysis with respect to $p$ in the proof of Proposition \ref{noncon-lemma-1} still applies here, thus we have the same bound.
\end{proof}

The only significant difference between the bound in Proposition \ref{noncon-compact} and the bound derived in Proposition \ref{noncon-lemma-1} is that the term with $\sigma^2_w$ is not multiplied by the step size $\eta_w$, thus appears to asymptotically behave as a constant. Therefore, in order to show that the right hand side in the above bound converges, we must treat $\sigma^2_w$ as a function of the number of stochastic gradients computed during each iteration. Recall that $\sigma^2_w$ is an upper bound on $\mathbb{E} \| \hat{g}_w - g_w^t \|_2^2$, and note from Lemma~\ref{lemma_var} that we can write it as $\sigma_w^2 = {\tilde{\sigma}_w^2}/{C}$, where $\tilde{\sigma}_w^2$ does not depend on $C$ or $T$, and $C$ is the number of sampled task instances used for each stochastic gradient computation, and each sampled task instance involves a constant number of function, gradient and Hessian evaluations. We can therefore define $C$ as an increasing function of $T$ in order for  $\sigma_w^2$ to decrease with $T$, while the total number of oracle evaluations performed by the algorithm will be $\mathcal{O}(CT)$.

To balance terms, we must choose $\eta_p$ and $\sigma_w^2$ to be of the same order with respect to $T$. Thus for some $\beta \in (0,1)$, let $\eta_p = \mathcal{O}(T^{-\beta})$ and $C = \mathcal{O}(T^{\beta})$. Since here $C$ grows with $T$, we can assume without loss of generality that $C > n$ (since if this were not the case, the only way we would get improvement over the 1/5 rate, to 1/4, would require $\beta =1$, which would mean C = m = n = T, which is not realistic). In this case, $\hat{G}_p^2$ can be numerically upper bounded as
\begin{equation}
\hat{G}_p^2 \coloneqq \frac{n(n+C-1)}{C} \hat{B}^2 = (\frac{n^2}{C} + n - \frac{n}{C})\hat{B}^2 \leq 2n \hat{B}^2
\end{equation}
Replacing $\hat{G}_p^2$ with this upper bound in the results from Proposition \ref{noncon-compact} and plugging in the appropriate step sizes completes the proof of Theorem \ref{noncon-thrm3}.

\section{Generalization Results}

\subsection{Proof of Proposition \ref{gen1}}
The result is a standard Rademacher complexity bound, see for example \citep{mohri2018foundations}, thus we omit the proof.

\subsection{Proof of Theorem \ref{gen_bound}}
\begin{proof}
Since $\mathcal{D}_{n+1}^K \times \mathcal{D}_{n+1}^J$ is a mixture distribution, we have, for any $w$,
\begin{align}
F_{n+1}(w) &= \mathbb{E}_{(D_{n+1, j}^{\text{train}}, D_{n+1, j}^{\text{test}}) \sim \mathcal{D}_{n+1}} [\hat{f}_{n+1, j}(w - \alpha \nabla \hat{f}_{n+1, j}(w,D_{n+1,j}^{\text{train}}), D_{n+1,j}^{\text{test}})]  \\
&= \sum_{i=1}^n a_i \mathbb{E}_{(D_{n+1,j}^{\text{train}}, D_{n+1,j}^{\text{test}}) \sim \mathcal{D}_{i}}  [ \hat{f}_{n+1,j}(w - \alpha \nabla \hat{f}_{n+1,j}(w,D_{n+1,j}^{\text{train}}), D_{n+1,j}^{\text{test}})  ] \\
&= \sum_{i=1}^n a_i F_i(w)
\end{align}
Therefore, using Proposition \ref{gen1} and a union bound over the $n$ tasks, we have that with probability at least $1-n \delta'$ over the choice of samples used to compute $\hat{F}_i(w)$,
\begin{align}
F_{n+1}(w^\ast) &=  \sum_{i=1}^n a_i F_i(w^\ast) \leq \sum_{i=1}^n a_i \hat{F}_i(w^\ast) + 2a_i\mathfrak{R}^i_{m_i}(\mathcal{F}) + a_i \hat{B} \sqrt{\frac{\log 1/\delta'}{2m_i} } 
\end{align}
Making the substitution $\delta = n \delta'$ and using the fact that $a_i \in \Delta_n$ and the definition of $w^\ast$ yields that
\begin{align}
F_{n+1}(w^\ast) &\leq \max_{p \in \Delta_n}  \sum_{i=1}^n p_i \hat{F}_i(w^\ast) + 2a_i\mathfrak{R}^i_{m_i}(\mathcal{F}) + a_i \hat{B} \sqrt{\frac{\log (n/\delta)}{2m_i} }  \\
&= \min_{w \in \mathcal{W}}  \max_{p \in \Delta_n}  \sum_{i=1}^n p_i \hat{F}_i(w^\ast) + 2a_i\mathfrak{R}^i_{m_i}(\mathcal{F}) + a_i \hat{B} \sqrt{\frac{\log (n/\delta)}{2m_i} }  \\
\end{align}
with probability at least $1-\delta$, which completes the proof.
\end{proof}





\section{Additional Experimental Results and Details} \label{app:experiments}

We performed all experiments on a 3.7GHz, 6-core Intel Corp i7-8700K CPU. For all experiments, there was no significant difference in the time required to run TR-MAML compared to MAML.

\subsection{Sinusoid Regression}

For the sinusoid regression experiments, we adapted the codebase from the original MAML paper \citep{finn2017model} available at \url{https://github.com/cbfinn/maml}, which is written in in Tensorflow \url{https://www.tensorflow.org/}. We used a batch size of 25 task instances with $J$ (the number of evaluation points in each task instance/few-shot learning episode) equal to $K$. We set $\eta_w = 10^{-3}$, $\alpha=10^{-3}$, and used one step of SGD update and the Adam optimizer for the meta-learning update step for $w$ for both TR-MAML and MAML, consistent with the original sinusoid experiments \citep{finn2017model}.  To update $p$ in TR-MAML, we used vanilla projected SGD (without an optimizer) with learning rate $\eta_p = 0.0001$ when $K=5$ and $\eta_p = 0.0002$ when $K=10$.

\subsection{Few-shot Image Classification}

\begin{table} 
  \caption{Omniglot $N$-way, $K$-shot classification accuracies (\%). After meta-training, 5,000 few-shot classification problems (task instances) are sampled uniformly from the 25 alphabets (tasks) used for meta-training, likewise for the 20 new meta-testing alphabets. For each alphabet, the average accuracy on task instances from that alphabet is computed, and statistics are taken across these average accuracies. `Weighted Mean' weighs the alphabet accuracies by the meta-training distribution, which corresponds to the quantity MAML aims to optimize, whereas `Mean' weighs all alphabets equally. `Worst' is the minimum alphabet accuracy, and `Std. Dev.' is the standard deviation across the alphabet accuracies, with 95\% confidence intervals given over three full runs for all statistics.}
  \label{omni-table_app0}
  \centering
  \resizebox{\columnwidth}{!}{%
  \begin{tabular}{llllllll}
    \toprule
   \multicolumn{2}{c}{}  & \multicolumn{3}{c}{Meta-training Alphabets}  &  \multicolumn{3}{c}{Meta-testing Alphabets}               \\
    \cmidrule(r){3-5}
    \cmidrule(r){6-8}
      $(N,K)$   &  Algorithm & Weighted Mean & Mean & Worst & Mean & Worst & Std. Dev. \\
    \midrule
        \multirow{2}{*}{(10,1)}  & MAML    & $ \mathbf{98.5 \pm 1.2} $ & $91.0\pm .4$ & $54.5 \pm  2.5$  & $\mathbf{85.9\pm 0.3}$  & $\mathbf{71.0 \pm 1.2}$ & $\mathbf{6.3 \pm .1}$  \\
             & TR-MAML  & $95.6 \pm .3$ &  $\mathbf{94.0\pm .1}$ & $ \mathbf{89.5 \pm 1.0 }$  & $83.6 \pm .5$ & $ 70.2 \pm 2.4$ & $6.6 \pm .3$     \\   
\cmidrule(r){1-8}
     \multirow{2}{*}{(10,5)}  & MAML  &$ \mathbf{99.1 \pm .1}$ & $95.0\pm .1$  & $70.1 \pm 2.8$  &$92.1 \pm .1$ & $82.9 \pm 0.1$ & $3.8 \pm .1$    \\
             & TR-MAML  & $98.5 \pm .4$ & $ \mathbf{98.6\pm  .4}$ & $\mathbf{96.2 \pm 1.0}$  & $\mathbf{93.8 \pm .7}$ & $\mathbf{87.7 \pm 1.4}$ &  $\mathbf{3.2 \pm .5}$     \\   
    \bottomrule
  \end{tabular}%
  }
\end{table}

For the image classification experiments on the Omniglot dataset, we adapted the codebase from the repository available at \url{https://github.com/AntreasAntoniou/HowToTrainYourMAMLPytorch} that implements in Pytorch \url{https://pytorch.org/} the experiments in the paper \citep{antoniou2018train}. Again we kept most of the default parameters consistent. The Adam optimizer was used for the meta-update of $w$ and vanilla SGD was used to update $p$. We set $\eta_p = 2.0 \times 10^{-5}$ for the 5-way experiments, $\eta_p = 1.6 \times 10^{-5}$ for the 10-way experiments, and $\eta_p = 1.0 \times 10^{-5}$ for the 20-way experiments. In all cases, we set $J=10$. After meta-training for 60,000 iterations with a batch size of 8, the most recent meta-trained model was evaluated on both the meta-testing and meta-training tasks (alphabets). One step of SGD was used for both meta-training and meta-testing in all experiments. Images were augmented by rotations of 90 degrees, with augmented images considered part of the same class (thus there were $20 \times 4 = 80$ images per class), but each image in each class in each task instance was rotated by the same amount. Additional results for the $10$-way classification case are shown in Table \ref{omni-table_app0}.

\bibliography{refs}
\bibliographystyle{plainnat}

\end{document}